\pdfoutput=1
\documentclass[11pt]{article}

\usepackage{style/acl}

\usepackage{times}
\usepackage{latexsym}

\usepackage[T1]{fontenc}
\usepackage[utf8]{inputenc}

\usepackage{microtype}

\usepackage{graphicx}
\usepackage{booktabs}
\usepackage{multirow}
\usepackage{threeparttable}
\usepackage{makecell}
\usepackage{array}
\usepackage{cellspace}
\usepackage{enumitem}
\usepackage{arydshln}
\usepackage{listings}
\usepackage{lipsum}
\usepackage{algorithm,algpseudocode}
\usepackage{bbm}
\usepackage{cleveref}
\usepackage[frozencache=true,cachedir=minted-cache]{minted}
\usepackage{arydshln}

\crefname{section}{\S\!}{\S\S}

\title{Leveraging Code to Improve In-context Learning for Semantic Parsing}

\newcommand{\authorsep}{\hspace{4ex}}
\newcommand{\instsep}{\hspace{4ex}}

\author{
  Ben Bogin\textsuperscript{1}\thanks{~~~Equal contribution} \authorsep
  Shivanshu Gupta\textsuperscript{2}$^{*}$ \authorsep
  Peter Clark\textsuperscript{1} \authorsep
  Ashish Sabharwal\textsuperscript{1} \\ 
  \hspace{1ex} \\
  \textsuperscript{1}Allen Institute for AI \instsep
  \textsuperscript{2}University of California Irvine  \\
  \texttt{\makecell{\{benb,peterc,ashishs\}@allenai.org, shivag5@uci.edu\\
}}
}

\makeatletter
\renewenvironment{minted@colorbg}[1]
 {\def\minted@bgcol{#1}%
  \noindent
  \begin{lrbox}{\minted@bgbox}
  \begin{minipage}{\linewidth-2\fboxsep}}
 {\end{minipage}%
  \end{lrbox}%
  \setlength{\topsep}{\bigskipamount}% set the vertical space
  \trivlist\item\relax % ensure going to a new line
  \colorbox{\minted@bgcol}{\usebox{\minted@bgbox}}%
  \endtrivlist % close the trivlist
 }
\makeatother

\begin{document}
\maketitle

\newif\ifdebug
% \debugtrue

\newif\ifcomments
% Uncomment line below to keep comments; comment line below to make them regular text
\commentstrue
\ifcomments
    \providecommand{\changed}[1]{{\protect\color{red}{#1}}}
    \providecommand{\bb}[1]{{\protect\color{olive}{[\textbf{BB}: #1]}}}
    \providecommand{\sg}[1]{{\protect\color{cyan}{[\textbf{SG}: #1]}}}
    \providecommand{\pc}[1]{{\protect\color{maroon}{[\textbf{PC}: #1]}}}
    \providecommand{\as}[1]{{\protect\color{blue}{[\textbf{AS}: #1]}}}
\else
    \providecommand{\changed}[1]{}
    \providecommand{\bb}[1]{}
    \providecommand{\sg}[1]{}
    \providecommand{\pc}[1]{}
    \providecommand{\as}[1]{}
\fi

\setminted{fontsize=\scriptsize, linenos}

\newcommand{\tightparagraph}[1]{\smallbreak\noindent\textbf{#1}}

\newenvironment{resultstable}[1][]{
    \begingroup
    \setlength{\tabcolsep}{3pt} % Default value: 6pt
    \begin{table*}{#1}
    \centering
    \small
}{
    \end{table*}
    \endgroup
}

\newenvironment{resultstablesinglecol}{
    \begingroup
    \setlength{\tabcolsep}{3pt} % Default value: 6pt
    \begin{table}
    \centering
    \small
}{
    \end{table}
    \endgroup
}

\let\spacedparagraph\paragraph

\definecolor{codegreen}{rgb}{0,0.6,0}
\definecolor{codegray}{rgb}{0.5,0.5,0.5}
\definecolor{codepurple}{rgb}{0.58,0,0.82}
\definecolor{backcolour}{rgb}{0.95,0.95,0.92}
\lstdefinestyle{mystyle}{
    backgroundcolor=\color{backcolour},
    commentstyle=\color{codegreen},
    keywordstyle=\color{magenta},
    numberstyle=\tiny\color{codegray},
    stringstyle=\color{codepurple},
    basicstyle=\ttfamily\scriptsize,
    breakatwhitespace=false,
    breaklines=true,
    captionpos=b,
    keepspaces=true,
    numbers=left,
    numbersep=5pt,
    showspaces=false,
    showstringspaces=false,
    showtabs=false,
    tabsize=2,
}
\lstset{style=mystyle}
\lstdefinelanguage{JavaScript}{
  morekeywords=[1]{break, continue, delete, else, for, function, if, in,
    new, return, this, typeof, var, void, while, with},
  % Literals, primitive types, and reference types.
  morekeywords=[2]{false, null, true, boolean, number, undefined,
    Array, Boolean, Date, Math, Number, String, Object},
  % Built-ins.
  morekeywords=[3]{eval, parseInt, parseFloat, escape, unescape},
  sensitive,
  morecomment=[s]{/*}{*/},
  morecomment=[l]//,
  morecomment=[s]{/**}{*/}, % JavaDoc style comments
  morestring=[b]',
  morestring=[b]"
}[keywords, comments, strings]
\lstdefinelanguage{Scala}{
  morekeywords={abstract,case,catch,class,def,%
    do,else,extends,false,final,finally,%
    for,if,implicit,import,match,mixin,%
    new,null,object,override,package,%
    private,protected,requires,return,sealed,%
    super,this,throw,trait,true,try,%
    type,val,var,while,with,yield},
  otherkeywords={=>,<-,<\%,<:,>:,\#,@},
  sensitive=true,
  morecomment=[l]{//},
  morecomment=[n]{/*}{*/},
  morestring=[b]",
  morestring=[b]',
  morestring=[b]"""
}
\setminted{bgcolor=backcolour,frame=leftline,numbersep=6pt,xleftmargin=15pt}

\begin{abstract}

In-context learning (ICL) is an appealing approach for semantic parsing due to its few-shot nature and improved generalization. However, learning to parse 
to rare domain-specific languages (DSLs) from just a few demonstrations is challenging, limiting the performance of even the most capable LLMs.

In this work, we show how pre-existing coding abilities of LLMs can be leveraged for semantic parsing by
(1) using general-purpose programming languages such as Python instead of DSLs and (2) augmenting prompts with a structured domain description that includes, e.g., the available classes and functions. We show that both these changes significantly improve accuracy across three popular datasets; combined, they lead to dramatic improvements (e.g., 7.9\% to 66.5\% on SMCalFlow compositional split) and can substantially improve compositional generalization, nearly closing the performance gap between easier i.i.d.\ and harder compositional splits.
Finally, comparisons across multiple PLs and DSL variations suggest that the similarity of a target language to general-purpose code is more important than prevalence in pretraining corpora.
Our findings provide an improved methodology for building semantic parsers in the modern context of ICL with LLMs.\footnote{\url{https://github.com/allenai/code-semparse}}

\end{abstract}
\section{Introduction}

\begin{figure}
  \centering
  \includegraphics[width=0.9\linewidth]{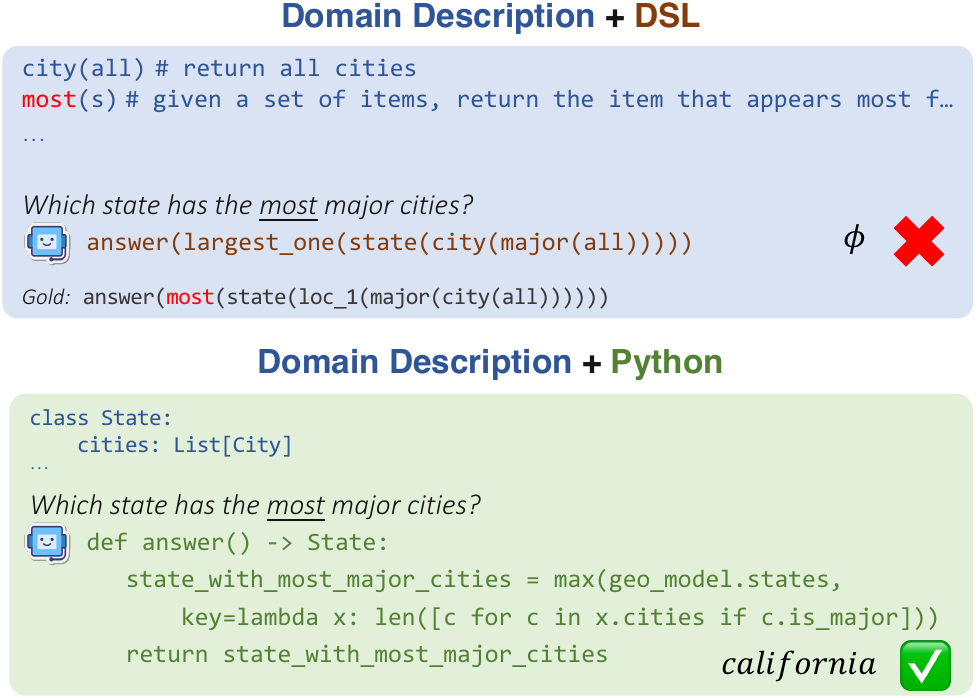}
  \caption{
  An example illustrating how moving the problem space from a DSL to a general-purpose programming language such as Python can improve output accuracy. When prompted with a DSL, the model doesn't use the operator \texttt{most}, resulting in an incorrect program. When prompted with Python, the model leverages its pre-existing knowledge of coding to produce the correct program and answer.}
  \label{fig:intro}
\end{figure}

Semantic parsing, the task of translating natural language utterances to structured meaning representations \cite{Zelle1996LearningTP,Kate2005LearningTT}
is a core requirement for 
building task-oriented dialog systems and voice assistants. % such as Alexa and Google Assistant 
This task is primarily addressed with two approaches: fine-tuning models on labeled datasets of utterances mapped to domain-specific language (DSL) programs \citep{xu-etal-2020-autoqa,oren-etal-2021-finding,gupta-etal-2022-structurally,yin-etal-2022-ingredients} and employing in-context learning (ICL; \citealp{Brown2020LanguageMA}) to prompt a large language model (LLM) with a few demonstrations. 

However, both strategies present significant limitations. Fine-tuning requires substantial pools of labeled data, which can be expensive and time-consuming to obtain. Crucially, fine-tuned models also struggle to compositionally generalize, e.g., to decode programs longer than seen during training or to emit unseen structures \citep{kim-linzen-2020-cogs,keysers2020measuring,bogin-etal-2022-unobserved,yao-koller-2022-structural}.
While ICL can improve compositional generalization in some cases \cite{anil2022exploring,qiu-etal-2022-evaluating,drozdov2023compositional,hosseini-etal-2022-compositional}, learning from a few demonstrations is challenging: LLMs need to not only understand the meaning of the input utterance but also learn how to correctly use a typically rare domain-specific language (DSL), given only few demonstrations.
This makes ICL sensitive to demonstration selection \cite{pmlr-v139-zhao21c}, which may not cover all functionalities and subtleties of a DSL. While prior work has tried to alleviate this with a better selection of demonstration \cite{Liu2021WhatMG,levy-etal-2023-diverse,gupta2023coveragebased}, such approaches require access to a large pool of labeled demonstrations to select from and are not applicable in a \textit{true} few-shot settings.

%\footnote{As opposed to the demonstration selection setup, which assumes access to a pool of examples and thus not truly few-shot.}

Given that LLMs show remarkable coding abilities in general-purpose programming languages (PLs; \citealt{chen2021evaluating,10.1145/3520312.3534862}), in this work, we ask two main questions: (1) How can we leverage these abilities to improve ICL-based semantic parsing? (2) Can LLMs compositionally generalize better with PLs rather than DSLs?

To investigate this, first, we replace DSLs with equivalent code written in \emph{popular programming languages} such as Python or Javascript. This helps better align the output space with pretraining corpora, obviating the need for LLMs to learn new
syntax, basic operations, or other coding practices from scratch. For example, consider Figure~\ref{fig:intro}: to select a state that has the \textit{most} major cities, an LLM prompted with a DSL needs to use the operator \texttt{most}, for which it might not be given an example. In contrast, with Python, the LLM can leverage its pre-existing knowledge of code to find such a state.

Second, we augment the ICL prompt with a structured description of the output meaning representation, which we refer to as \emph{Domain Description (DD)}. This provides domain-specific information such as types 
of entities, attributes, and methods (e.g., \texttt{State} and its attributes in Figure~\ref{fig:intro}). While such descriptions can also be added to DSLs, we find that domain descriptions for PLs are easier to precisely define with explicit declarations of objects, methods, their signatures, etc. Furthermore, LLMs are more likely to leverage descriptions with PLs rather than DSLs, as using previously defined objects and methods is a common coding practice.

We evaluate our approach on both ChatGPT\footnote{\label{footnote:chatgpt}\url{https://chat.openai.com/}} and the open-source Starcoder model \cite{li2023starcoder}, by implementing Python-executable environments for three complex semantic parsing benchmarks, namely GeoQuery \citep{Zelle1996LearningTP}, Overnight \citep{wang-etal-2015-building}, and SMCalFlow \citep{andreas-etal-2020-task}, and annotating them with Python programs and DDs.

In a true few-shot setting, where only a few (e.g., 10) labeled examples are available to use as demonstrations, we find that PL prompts with DDs dramatically improve execution-based accuracy across the board, e.g., 49.7 points absolute improvement (31.0\% to 80.7\%) on the length split of GeoQuery, compared to the standard ICL approach of a DSL-based prompt with no DD.
Prompting a model with Python and domain description can often even eliminate the need for many demonstrations: with just a \textit{single} demonstration, accuracy on a compositional split of GeoQuery reaches 80\%, compared to 17\% for DSL prompting with no DD. In fact, for two datasets, a single PL demonstration with DD outperforms DSL prompts with as many as 25 demonstrations and an equivalent DD. 
Interestingly, we find that employing Python with a DD substantially improves compositional generalization, almost entirely closing the compositionality gap, i.e., the performance difference between an i.i.d.\ split and harder compositional splits. 

One might hypothesize that the strong performance of Python is due to its prevalence in the pretraining corpus \cite{cassano2023knowledge}. To investigate this, we evaluate the performance of PLs whose popularity differs from that of Python.
Surprisingly, we find that prevalence in pretraining corpora does not explain superiority: both Scala, a PL much rarer than Python, and Javascript, which is much more prevalent, perform roughly similarly. SQL, a common query language, performs better than DSLs, but worse than the other more general-purpose PLs. Further analyses with simplified versions of DSLs indicate that even rare DSLs, as long as they resemble general-purpose code, might perform nearly as well as PLs, provided a detailed DD is used.

In conclusion, we demonstrate that using popular PLs instead of DSLs and adding domain descriptions dramatically improves ICL for semantic parsing while nearly closing the compositionality gap. Further, we show that
when LLMs are used for semantic parsing, it is better to either prompt them with PLs or design DSLs to resemble popular PLs. Overall, these findings suggest an improved way of building semantic parsing applications in the modern context of in-context learning with LLMs.

\newcommand{\code}[1]{\texttt{\scriptsize{#1}}}
\newcommand{\rowSpace}{2ex} % Define a new command for the row space
\newcommand{\myhline}{\noalign{\vskip 1mm}\hline\noalign{\vskip 1mm}}
\newcommand{\reducedRowSpace}{0.1ex} % Define a new command for the reduced row space

\begin{table*}[!ht]
\setlength\cellspacetoplimit{2pt}
\setlength\cellspacebottomlimit{2pt}
\setlength{\tabcolsep}{3pt} % Adjusts the space between columns
\centering
\scriptsize
\begin{tabular}{Sc Sc Sc Sc Sl}
\toprule
\textbf{Dataset} & \textbf{MR} & \textbf{\# chars} & \textbf{Depth} & \textbf{Example}\\
\midrule
        GeoQuery  & \textit{input}  & & & \textit{How high is the highest point in the largest state?}  \\[\reducedRowSpace]
          & FunQL  & 49.4 & 4.8 &\makecell[l]{\code{\makecell[l]{answer(elevation\_1(highest(place(loc\_2(largest(state(all)))))))}}}  \\[\reducedRowSpace]
          & Python & 115.4 & &\makecell[l]{\code{\makecell[l]{largest\_state = max(geo\_model.states, key=lambda x: x.size) \\
    return largest\_state.high\_point.elevation}}} \\ \myhline

        Overnight  & \textit{input}  &  & & \textit{person whose gender is male and whose birthdate is 2004}  \\[\reducedRowSpace]
          & $\lambda$-DCS  & 282.0 & 6.8 & \makecell[l]{\code{\makecell[l]{(call SW.listValue (call SW.filter (call SW.filter (call SW.getProperty (call\\SW.singleton en.person) (string !type)) (string gender) (string =) \\en.gender.male) (string birthdate) (string =) (date 2004 -1 -1)))}}}  \\ 
        
          & \makecell{$\lambda$-DCS\\(Simp.)}  & 164.1 & 6.0 & \makecell[l]{\code{\makecell[l]{(listValue (filter (filter (getProperty en.person !type) \\gender = en.gender.male) birthdate = 2004))}}} \\ 
        
          & Python & 270.0 & & \makecell[l]{\code{\makecell[l]{people\_born\_in\_2004 = [p for p in api.people if p.birthdate == 2004]\\
    males\_born\_in\_2004 = [p for p in people\_born\_in\_2004 if p.gender == \\ ~~~~~~~~~~~~~~~~~~~~~~Gender.male]\\
    return males\_born\_in\_2004}}} \\ \myhline

        SMCalFlow  & \textit{input}  &  & & \textit{Make an appointment in Central Park on Friday.}  \\[\reducedRowSpace]
        
          & Dataflow  & 372.6 & 8.7 & \makecell[l]{ \code{\makecell[l]{(Yield :output (CreateCommitEventWrapper :event (CreatePreflightEventWrapper \\ :constraint (Constraint[Event] :location ( ?= \# (LocationKeyphrase "Central\\ Park")) :start (Constraint[DateTime] :date ( ?= (NextDOW :dow \# (DayOfWeek\\ "FRIDAY"))))))))}}}  \\[\reducedRowSpace]
        
          & \makecell{Dataflow\\(Simp.)} & 118.7 & 4.2 &  \makecell[l]{\code{\makecell[l]{CreateEvent( AND( at\_location( Central Park ) , starts\_at( NextDOW( \\FRIDAY ) ) ) )}}} \\[\reducedRowSpace]
        
          & Python & 174.4 & & \makecell[l]{ \code{\makecell[l]{api.add\_event(Event(subject="Appointment in Central Park", starts\_at=[\\DateTimeClause.get\_next\_dow(day\_of\_week="Friday")], location="Central Park"))}}} \\ \myhline
\end{tabular}
    \caption{Sample input, program, average program length and average maximum depth for each dataset and meaning representation considered. Depth is computed based on parentheses.}
    \label{tab:datasets}
\end{table*}

\section{Related Work}

\paragraph{Compositional Generalization.} Semantic parsing has been studied extensively in recent years in the context of compositional generalization (CG), where models are evaluated on examples that contain unseen compositions of structures, rather than easier i.i.d.\ train-test splits \cite{finegan-dollak-etal-2018-improving}. Initial work on CG focused on fine-tuning-based approaches. As simply scaling the model size or amount of data CG has been shown to be insufficient for improving CG
\cite{hosseini-etal-2022-compositional,qiu-etal-2022-evaluating}, prior work explored approaches like specialized architectures, \cite{herzig-berant-2021-span,bogin-etal-2021-latent,yin-etal-2021-compositional,lindemann-etal-2023-compositional-generalization}, data augmentation \cite{andreas-2020-good,qiu-etal-2022-improving,akyurek2021learning}, training data selection \cite{bogin-etal-2022-unobserved,gupta-etal-2022-structurally}, etc.
With the increasing prevalence of in-context learning with LLMs, recent works have focused on improving its compositional generalization through better demonstration selection \cite{Liu2021WhatMG,ye2023ceil,an-etal-2023-context,li-etal-2023-unified,zhang-etal-2022-active,su2023selective,levy-etal-2023-diverse,gupta2023coveragebased,gupta2024gistscore}.
% Various works have shown how generalization can be improved by selecting better demonstrations \cite{Liu2021WhatMG,ye2023ceil,an-etal-2023-context,li-etal-2023-unified,zhang-etal-2022-active,su2023selective} or covering a diverse set of program structures \cite{bogin-etal-2022-unobserved,gupta-etal-2022-structurally,levy-etal-2023-diverse,gupta2023coveragebased}, 
However, all these methods require a large pool of demonstrations or annotation efforts. In contrast, we show that by leveraging pre-existing coding abilities, LLMs do not need as many examples to generalize.

\paragraph{Effect of Meaning Representations.} To address specific challenges with DSLs, previous work has proposed to work with simpler meaning representations (MRs) \cite{herzig2021unlocking,li-etal-2022-exploring-secrets,Wu-etal:2023:ReCOGS} or synthetic NL utterances \cite{shin-etal-2021-constrained}, or prompting models with the grammar of the DSL \cite{wang2023grammar}. Recently, \citet{jhamtani2023natural} used Python to satisfy virtual assistant requests. Differently from that, our work provides an extensive study exploring the advantage of using code and domain descriptions in semantic parsing, across different datasets and PLs.

\paragraph{Code Prompting.} Numerous works have shown that code-pretrained LLMs can be leveraged to improve various tasks such as arithmetic reasoning, commonsense reasoning, and others with prompts that involve code \cite{gao2022pal,madaan-etal-2022-language,Chen2022ProgramOT,zhang-etal-2023-exploring,hsieh2023tool}.
In this work, we show for the first time how to effectively use code prompts for semantic parsing, 
demonstrating that when the output of the task is already programmatic and structured, performance gains can be dramatically high.

\section{Setup}
\label{sec:setup}
Given a natural language request $x$ and an environment $e$, our task is to ``satisfy'' the request as follows by executing a program $z$:
If $x$ is an \emph{information-seeking question},
program $z$ should output
the correct answer $y$; if $x$ is an \emph{action request},
$z$ should update the environment $e$ appropriately.
For example, an information-seeking question could be \textit{``what is the longest river?''}, where $e$ contains a list of facts about rivers and lengths, and the answer $y$ should be returned based on these facts. An action request could be \textit{``set a meeting with John at 10am''} where $e$ contains a database with a list of all calendar items, and the task is to update $e$ such that the requested meeting event is created. The environment $e$ can be any type of database or system that provides a way to retrieve information or update it using \textit{a formal language program}. Each environment accepts a different formal language for $z$, and has its own specific list of accepted operators (see Table~\ref{tab:datasets} for examples of different formalisms used in this work).

We focus on the \textbf{true few-shot setup} where only a small ($\leq 10$) set of demonstrations is used. Specifically, we assume knowledge of the formalism and operators supported by $e$, and a set of training examples $\{(x_i,z_i)\}_{i=1}^k$ where $k$ is small ($\leq 10$), and no other data, labeled or otherwise.

\section{Domain-Augmented PL Prompts}
\label{sec:python}

Semantic parsing studies have traditionally used DSLs. We posit that using general-purpose PLs with a structured description of the domain in hand could better exploit the potential of modern LLMs, which are pretrained on a mix of code and natural language. % Next, we elaborate on both these aspects.

\begin{figure}
  \centering
  \includegraphics[width=0.9\linewidth]{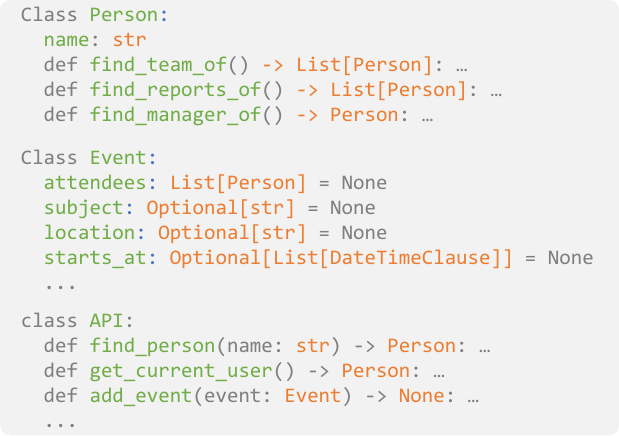}
  \caption{A partial example of a domain description containing the names of all objects and operators (in green) and type signatures (in orange).}
  \label{fig:python-prompt}
\end{figure}

\paragraph{Leveraging Existing Coding Knowledge.}
While DSLs tailored to specific domains can be valuable for trained domain experts, their rarity makes it challenging for LLMs to learn them from just a few demonstrations. In contrast, PLs are prevalent in pretraining corpora; by prompting LLMs to generate PLs rather than DSLs, LLMs can leverage their existing coding knowledge without the need to learn the syntax and standard operations for a new language from scratch.

For instance, consider the operator \texttt{most} in Figure~\ref{fig:intro}. LLMs with no prior knowledge of the given DSL struggle to correctly apply this operator without sufficient demonstrations. However, with Python, the model can exploit its parametric knowledge to perform this operation by employing the built-in \texttt{max} and \texttt{len} operators of Python, along with list comprehension. Another example is filtering sets of items in $\lambda$-DCS (Table~\ref{tab:datasets}, Overnight). Using a rare DSL, models must learn how to correctly use the filter operator from just a few demonstrations. However, LLMs have likely already seen a myriad of filtering examples during pretraining, e.g. in the form of Python's conditional list comprehension.

\paragraph{Domain Descriptions.}
While using PLs allows the model to leverage its parametric knowledge of the language's generic aspects, the LLM is still tasked with understanding domain-specific functionalities from a few in-context demonstrations. This is challenging, often even impossible, in a true few-shot setup, where the few fixed demonstrations may not cover all the functionality necessary to satisfy the test input request. A line of prior work alleviated this issue by selecting the most relevant demonstrations for every test input \cite{levy-etal-2023-diverse,gupta2023coveragebased}, but this approach typically requires a large labeled pool of demonstrations.

To address this challenge in a true few-shot setup, we propose an intuitive solution that naturally aligns with the use of PLs: providing the model with a \emph{Domain Description} (DD) outlining the available operators. Specifically, when using PLs, we prefix the ICL prompt with definitions of the domain classes, methods, attributes, and constants exactly as they are defined in the environment, with the implementations of specific methods concealed for prompt brevity (e.g., replaced with `\texttt{...}' in Python).

Figure~\ref{fig:python-prompt} provides a snippet of the Python DD for 
% the python environment of 
SMCalFlow \cite{andreas-etal-2020-task}, where users can create calendar events with certain people from their organization. Perhaps most importantly, DDs include the names of all available operators (highlighted in green in the figure). Without a list of available operators and relevant demonstrations, models are unlikely to generate a correct program. The type signatures (highlighted in orange in the figure) provide additional important information on how these operators and attributes can be used. The complete DDs are deferred to App.~\ref{app:dd}.

While DDs can also be used with DSLs, there's typically no consistent and formal way to write such descriptions. In contrast, DDs for PLs are not only easier to write, they could be particularly effective as pretraining corpora contain countless examples of how previously defined classes and methods are used later in the code. As we will empirically demonstrate in Section \ref{sec:experiments}, DDs are indeed utilized more effectively with PLs than with DSLs.

\paragraph{Prompt Construction.} The prompt that we use is a concatenation of the domain description (such as the example in \Cref{fig:python-prompt}) and demonstrations (such as the inputs and MRs in \Cref{tab:datasets}) for a given environment. See App.~\ref{app:prompt} for the exact format.

\section{Experimental Setup}
\label{subsec:experimental-setup}

\subsection{Datasets and Environments}

\paragraph{Datasets.} We experiment with three semantic parsing datasets, covering both information-seeking questions and action requests. See \Cref{tab:datasets} for examples.
\begin{itemize}[leftmargin=*,itemsep=4pt,parsep=0pt,topsep=2pt]
    \item \textbf{GeoQuery} \cite{Zelle1996LearningTP} contains user utterances querying about geographical facts such as locations of rivers and capital cities. % We use the FunQL formalism \cite{kate-mooney-2006-using}, which is functional and variable-free.
    % \as{what's a good place to include dataset sizes? Here? As a column in \Cref{tab:datasets}? In splits paragraph? Need some sense of it to assess significance vs.\ noise}
    \item \textbf{SMCalFlow} \cite{andreas-etal-2020-task} contains user requests to a virtual assistant helping with actions such as setting up organizational calendar events. %, using the Dataflow formalism created specifically for the task.
    \item \textbf{Overnight} \cite{wang-etal-2015-building} contains queries about various domains; in this work, we use the `social network' domain, with questions about people's employment, education, and friends. % Overnight uses the $\lambda$-DCS formal language \cite{liang-etal-2011-learning}.
\end{itemize}

\paragraph{DSLs.} Unless mentioned otherwise, we experiment with the original DSLs of the tasks: FunQL \cite{kate-mooney-2006-using} for GeoQuery, Dataflow for SMCalFlow, and $\lambda$-DCS \cite{liang-etal-2011-learning} for Overnight. We also experiment with a simpler version of $\lambda$-DCS for Overnight, where we reversibly remove 
certain redundant keyword, and Dataflow-Simple \cite{meron2022simplifying}, a simpler (and less expressive) version of Dataflow, to better understand the effect of the design of DSLs (\S\ref{subsec:why-python}).

\paragraph{Dataset Splits.} For each dataset, we experiment with both i.i.d. splits (random splits between training and test sets) and compositional generalization splits, as detailed in App~\ref{app:dataset-splits}. All results are reported on development sets where available, except for \Cref{tab:results-GPT,tab:results-starcoder}, where we use the test sets.

\paragraph{Executable environments.}
As described in \S~\ref{sec:setup}, an environment is capable of executing a program $z$ and either outputting an answer $y$ (e.g., the name of a river) or modifying its own state (e.g., creating an event). In this work, for each dataset, we use an existing executable environment for the DSL formalism and implement one for Python. 
% We provide an overview of how we create the Python environments here, and refer to App.~\ref{app:envs} for implementation details of all of the environments we use.

To implement the Python environments, we analyze the original DSL programs to identify the requisite classes, their properties, and their methods,
% based on usage of the original DSL
and then write Python code to create an executable environment. Importantly, whenever possible, we retain the original names of properties and constants used in the DSLs, ensuring that performance improvements can be attributed to the change in MR rather than changes in naming. We refer to App.~\ref{app:envs} for implementation details of all of the environments we use.

\subsection{Evaluation}
\paragraph{Metrics.} The executable environments we have for all datasets, for both DSL and Python, allow us to compute \textit{execution-based} accuracy. For GeoQuery and Overnight, we compare answers returned by generated programs to those generated by gold DSL programs. For SMCalFlow, we compare the state (i.e., calendar events) of the environments after executing gold and predicted programs. For DSL experiments, we additionally provide Exact Match metric results in App.~\ref{app:additional_results}, which are computed by comparing the generated programs to gold-annotated programs.

We run all experiments with three seeds, each with a different sample of demonstrations, and report average accuracy. For each seed, the same set of demonstrations is used across different test instances, MRs, and prompt variations. Standard deviations for main results are provided in \Cref{app:results-std}.

\paragraph{Conversion to Python.} 
To generate Python programs demonstrations, we convert a subset of the DSL programs of each dataset to Python using semi-automatic methods while validating them by ensuring they execute correctly. See App.~\ref{app:annotations} for details.

\paragraph{Models.} We experiment with OpenAI's ChatGPT ({\small\texttt{gpt-3.5-turbo-0613}}) and the open-source StarCoder \cite{li2023starcoder}. Since GPT's maximum context length is longer, we conduct our experiments with GPT with $k=10$ demonstrations and provide main results for StarCoder with $k=5$. We use a temperature of 0 (greedy decoding) for generation.

\paragraph{Domain Descriptions for DSLs.}
For a thorough comparison, we also provide DDs for each DSL, containing similar information as the PL-based DDs (\S\ref{sec:python}). We manually write these DDs based on the existing environments, listing all operators and describing type signatures. We write the descriptions of operators in natural language (NL); For GeoQuery we also experiment with code descriptions, where names of operators are followed by Python-like signatures (see App.~\ref{app:dd} for all DDs).
Unless mentioned otherwise, Full DD for DSLs refers to the NL version.

We note that providing DDs for DSLs is often not as straightforward as for PLs; we design the DSL-based DDs to be as informative as possible but do not explore different description design choices. This highlights another advantage of using PLs---their DDs can simply comprise extracted definitions of different objects without the need to describe the language itself.

\section{Results}
\label{sec:experiments}

We first compare Python-based prompts with DSL-based prompts and the effect of DDs (\S\ref{subsec:python-vs-dsl}). We then experiment with several other PLs and variations of DSLs to better understand how the design of the output language affects performance (\S\ref{subsec:why-python}). 

\subsection{Python vs DSLs}
\label{subsec:python-vs-dsl}

\spacedparagraph{Baselines and Ablations.} We compare multiple variations of DDs. \textit{List of operators} simply lists all available operators without typing or function signatures (i.e., we keep only green text in Figure~\ref{fig:python-prompt}). \textit{Full DD} contains the entire domain description, while \textit{DD w/o typing} is the same as Full DD, except that it does not contain any type information (i.e., none of the orange text in Figure~\ref{fig:python-prompt}).

\begin{table}[!t]
\centering
\scriptsize
\setlength{\tabcolsep}{2pt}
\begin{threeparttable}
\begin{tabular}{@{}llcccccccc@{}}
\toprule
 & & \multicolumn{4}{c}{\textbf{GeoQuery}} & \multicolumn{2}{c}{\textbf{SMC-CS}} & \multicolumn{2}{c}{\textbf{Overnight}} \\
 \cmidrule[0.4pt](r{0.125em}){3-6}%
 \cmidrule[0.4pt](lr{0.125em}){7-8}%
 \cmidrule[0.4pt](lr{0.125em}){9-10}%
 & &  i.i.d. & Templ. & TMCD & Len. & i.i.d. & 0-C & i.i.d. & Templ. \\ 
\midrule
\multirow{3}{*}{\textbf{\rotatebox[origin=c]{90}{DSL}}} & No DD         & 37.6 & 34.2 & 43.5 & 31.0 & 42.9 & 7.3 & 34.0 & 23.1 \\
 & List of operators                      & 48.6 & 31.8 & 47.3 & 38.0 & 45.9 & 20.1 & 38.1 & 24.0 \\
 & Full DD (code)                      & 51.9 & 34.5 & 53.6 & 43.9 & - & - & - & - \\
 & Full DD (NL)                            & 61.0 & 41.5 & 52.5 & 39.4 & 40.7 & 20.7 & 38.8 & 23.2 \\
\midrule
\multirow{4}{*}{\textbf{\rotatebox[origin=c]{90}{Python}}} & No DD      & 72.6 & 54.6 & 67.1 & 57.3 & 58.0 & 28.0 & 62.8 & 69.2 \\
 & List of operators                      & 83.5 & 83.0 & 81.9 & 75.3 & 59.6 & 46.9 & 61.7 & 71.5 \\
 & DD w/o typing                      & 82.1 & 83.1 & 83.4 & 75.7 & 69.1 & 65.6 & 62.1 & 68.5 \\
 & Full DD                            & \textbf{84.4} & \textbf{84.4} & \textbf{84.9} & \textbf{80.7} & \textbf{69.2} & \textbf{66.7} & \textbf{64.5} & \textbf{72.9} \\
\bottomrule
\end{tabular}
\end{threeparttable}%
\caption{Execution accuracy of GPT-3.5-turbo, comparing Python-based prompts with DSL-based prompts across different DD variations, when 10 in-context demonstrations are used. Python-based prompts with Full DD consistently outperform DSL-based prompts by substantial amounts. Test sets results. 
% SMC-CS is SMCalFlow-CS.
}
\label{tab:results-GPT}
\end{table}

\paragraph{Main Results.} \Cref{tab:results-GPT} presents the results for ChatGPT ($k$=10), while \Cref{tab:results-starcoder} in App.~\ref{app:results-starcoder} shows results for Starcoder ($k$=5). We observe that Python programs without a DD outperform not only DSLs without a DD but even surpass DSLs prompted with a full DD across all splits for GPT and on most splits for Starcoder. Python with a full DD performs best in all 8 splits for GPT and on 5 splits for Starcoder. Notably, for ChatGPT, using Python with Full DD almost entirely eliminates the compositionality gap, i.e., the difference in performance between the i.i.d.\ split and compositional splits.

Ablating different parts of the DDs (rows ``List of operators'' and ``DD w/o typing'') reveals that in some cases, most of the performance gain for Python-based prompts is already achieved by adding the list of operators (e.g., GeoQuery i.i.d.\ split), while in other cases (e.g., GeoQuery length split) providing typing and signatures further improves accuracy. For DSL-based prompts, both formal DDs and natural language (rows Full DD formal/NL) underperform Python-based prompts, suggesting that Python's performance gains are not only due to descriptions being formal.

\begin{figure}
  \centering
  \includegraphics[width=\linewidth]{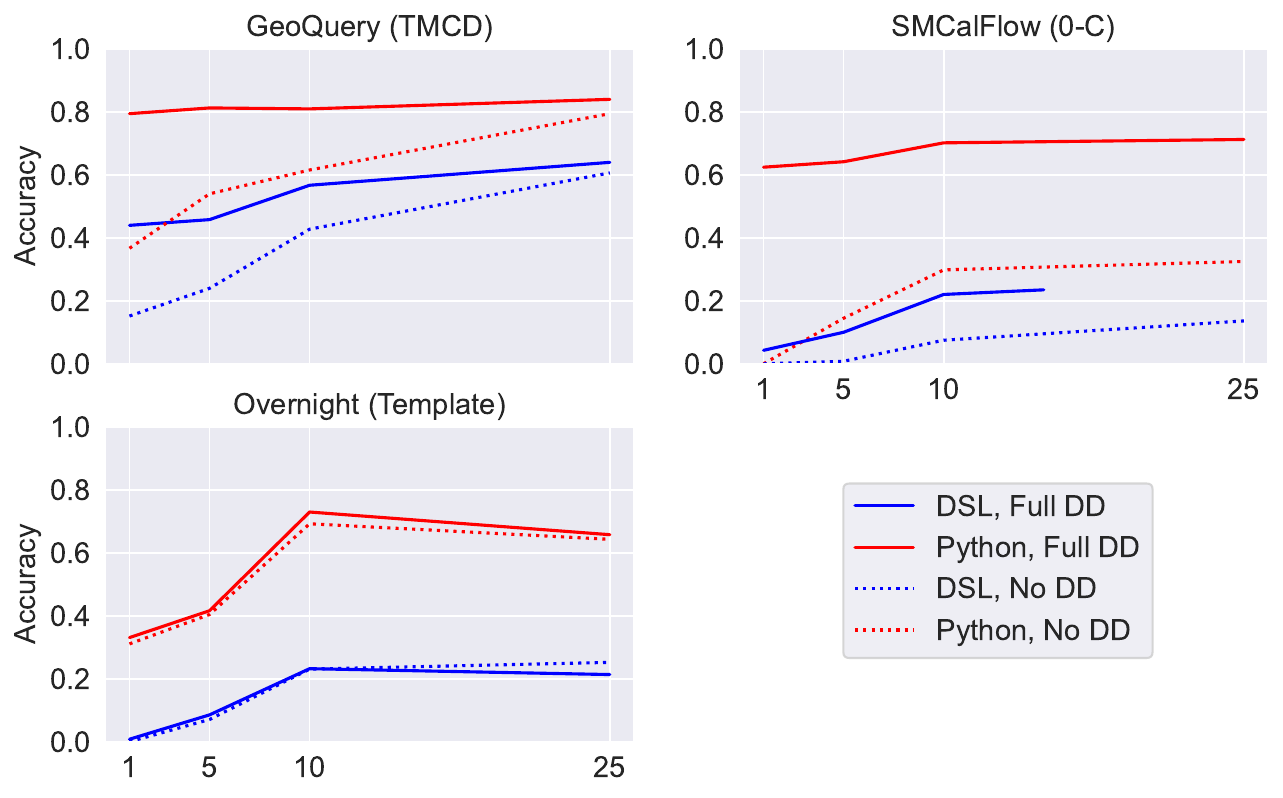}
  \caption{
  Execution accuracy for varying number of demonstrations. In almost all cases, Python outperforms DSL, both with a domain description and without, across different numbers of demonstrations (prompt for SMCalFlow, DSL, Full DD could not fit more than 15 examples given the model's context length limitation).
   }
  \label{fig:n-demonstrations-result}
\end{figure}

\paragraph{Prompt Length Trade-off.}
Figure~\ref{fig:n-demonstrations-result} demonstrates that using Python consistently outperforms DSLs across varying numbers ($k$) of demonstrations.
For both GeoQuery and SMCalFlow, just a single demonstration with a DD outperforms 25 demonstrations without a DD. However, the impact of DDs depends on the dataset and the domain: DDs lead to dramatic gains for the more complex SMCalFlow, but are less impactful in Overnight where the domain is small.
% , the impact of DDs is limited.

Considering a real-world setup with constrained resources, where one might want to optimize performance given a maximum prompt length, we also investigate accuracy as a function of the \emph{total number of prompt tokens} for three Python DD variations. We find that the optimal point in the trade-off between DD detail and number of demonstrations in the prompt varies per dataset (see Figure~\ref{fig:n-tokens-result} in App.~\ref{app:accuracy_vs_tokens}). For Overnight, where the domain is simple, using demonstrations alone might suffice. However, for both GeoQuery and SMCalFlow, having the Full DD is preferred whenever it can fit. 

\begin{figure}
  \centering
  \includegraphics[width=0.9\linewidth]{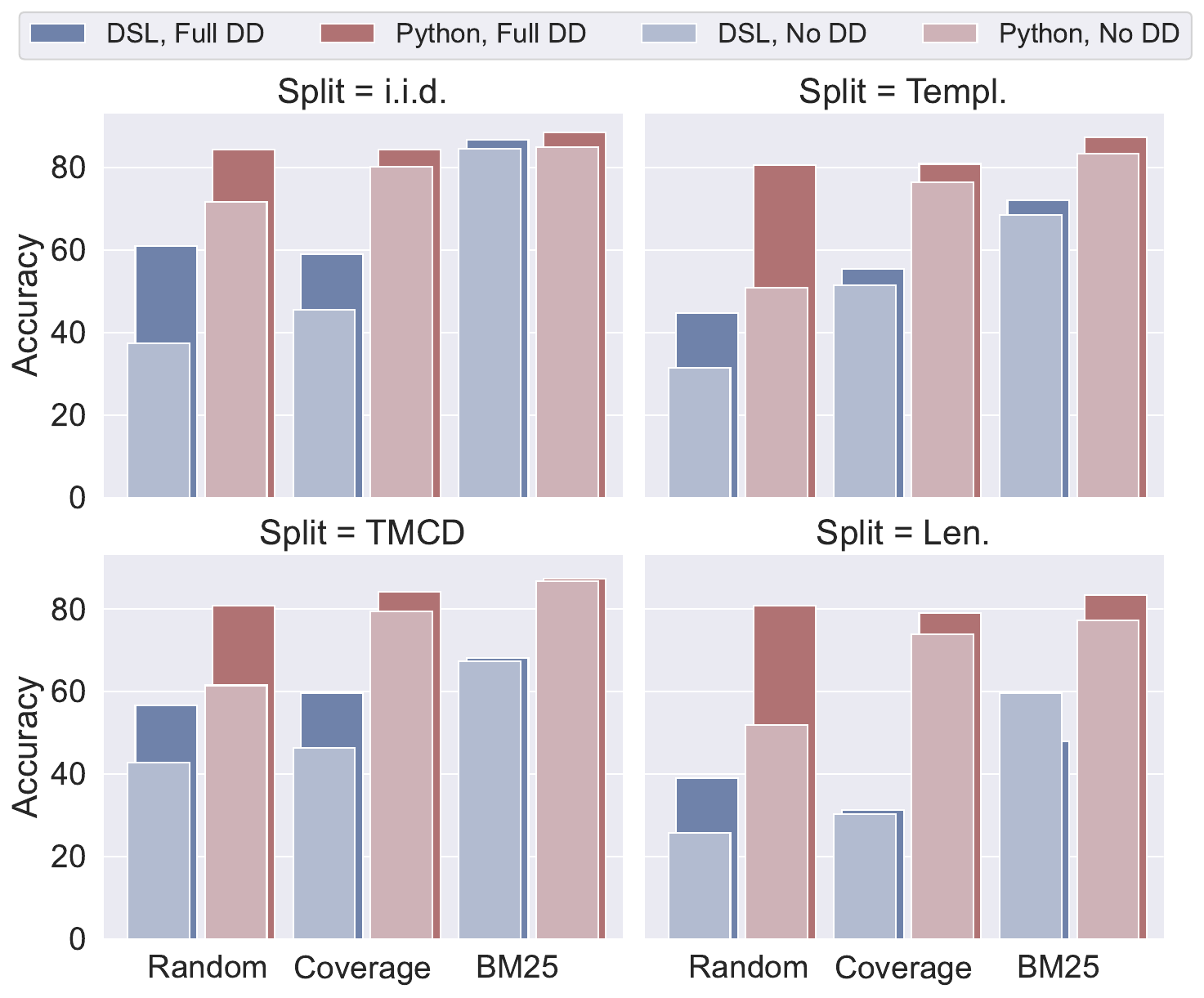}
  \caption{Python-based prompts, both with and without DD, consistently outperform DSL-based prompts, even with better demonstrations, for every split of GeoQuery.
  }
  \label{fig:pldsl}
\end{figure}

\paragraph{Effect of Better Demonstrations Selection.} Our results so far have demonstrated performance with a \emph{random}, fixed set of demonstrations, in line with our goal of minimizing labeling workload. 
However, in some scenarios, the budget may allow access to larger pools of demonstrations, in turn allowing more sophisticated demonstration selection methods to be applied. 
To evaluate our approach in such a setting, we additionally experiment with two selection methods.
% In the next experiment, we evaluate the performance of PL-based prompts when a large pool of labeled data is available, 
% allowing more sophisticated demonstration selection methods to be applied.

The first method optimizes for \textit{operator coverage} \cite{levy-etal-2023-diverse,gupta2023coveragebased}
by selecting a fixed set of demonstrations that cover as many of the operators as possible. This is achieved by greedily and iteratively selecting demonstrations to cover operators (see App.~\ref{app:demo_selection} for details). This fixed set covers 68\% to 81\% of the operators with $k=10$ (coverage varies across splits). Our second selection method is similarity-based retrieval: given a test example utterance, we retrieve the training examples with the most similar utterances using BM25 \cite{Robertson2009ThePR}.

We present the results for the different demonstration selection methods in Figure~\ref{fig:pldsl} for GeoQuery, for which we have annotated the entire training set with Python programs. We observe that for every selection method, both with and without DD, Python-based prompts consistently outperform DSL-based prompts.

\paragraph{Error Analysis.}

We now analyze the kinds of errors made by the LLM when prompted with Python and a DD. For SMCalFlow and ChatGPT, the development set of the compositional split (of size 250) resulted in 78 errors on one of the seeds; 
we include common examples of errors in App. \ref{app:examples} (Table \ref{tab:errors}, with examples of correct predictions in Table \ref{tab:successes}). % \as{moved previous $\leftarrow$ sentence here, before going into numbers.}
42 (54\%) of the errors were because the program failed to execute. The remaining 36 were due to incorrect execution. Closer analysis revealed most of these errors to be due to failure to understand the input utterance or not using the API correctly. A small fraction (11, 14\%) of the error instances were found to be unsupported by the original environment or our Python re-implementation. For GeoQuery, on the other hand, among the 18 errors made by ChatGPT on the development set of the TMCD split (of size 100) on one of the seeds, only 8 were attributed to model errors, while 8 were due to discrepancies in the dataset\footnote{These include incorrect FunQL annotation or discrepancies in the GeoBase database underlying GeoQuery, e.g., \textit{Mount McKinley} is referred to as “Mount McKinley” and “mckinley” leading to ambiguity.} and 2 resulted from environment limitations.

The above analysis suggests that while using PLs and DDs greatly improves the performance of LLMs, there is still scope for improvement in more complex domains (like SMCalFlow). Future work can explore how to ensure LLMs remain faithful to the DD and how to design PL environments to be more amenable to LLMs.

\begin{table}[!t]
\centering
\scriptsize
\setlength{\tabcolsep}{2pt}
\begin{threeparttable}
\begin{tabular}{@{}lllcccccccc@{}}
\toprule
  & & \multicolumn{4}{c}{\textbf{GeoQuery}} & \multicolumn{2}{c}{\textbf{SMC-CS}} & \multicolumn{2}{c}{\textbf{Overnight}} \\
 \cmidrule[0.4pt](r{0.125em}){3-6}%
 \cmidrule[0.4pt](lr{0.125em}){7-8}%
 \cmidrule[0.4pt](lr{0.125em}){9-10}%
 & & i.i.d. & Templ. & TMCD & Len. & i.i.d. & 0-C & i.i.d. & Templ. \\ 
\midrule

\multirow{2}{*}{DSL} & No DD    & 38.7 & 32.1 & 42.7 & 25.8 & 42.9 & 7.5 & 34.0 & 23.1 \\
 & Full DD                                  & 61.9 & 44.8 & 56.7 & 39.1 & 40.8 & 22.0 & 38.8 & 23.2 \\
\midrule
\multirow{2}{*}{Python} & No DD & 71.6 & 50.9 & 61.5 & 51.8 & 58.1 & 29.7 & 62.8 & 69.2 \\
 & Full DD                                  & 83.1 & 80.6 & 80.9 & 80.9 & \textbf{69.3} & 69.9 & 64.5 & \textbf{72.9} \\
\midrule
\multirow{2}{*}{Javascript} & No DD               & 80.0 & 72.1 & 75.2 & 73.6 & 64.1 & 46.1 & 61.6 & 51.3 \\
 & Full DD                                         & 81.1 & 80.0 & 77.3 & 73.6 & 68.1 & 71.6 & 61.1 & 50.1 \\
\midrule
\multirow{2}{*}{Scala} & No DD                    & 82.5 & 73.3 & 73.9 & 68.5 & 62.3 & 45.9 & 69.7 & 65.1 \\
 & Full DD                                         & \textbf{83.5} & \textbf{83.3} & \textbf{82.4} & \textbf{82.1} & 69.2 & \textbf{72.4} & \textbf{69.9} & 61.9 \\
\midrule
\multirow{2}{*}{SQL} & No DD                    &  65.7 & 56.6 & 61.8 & 45.6  &  &  &  &  \\
 & Full DD$^\dagger$                                         & 73.1 & 68.7 & 75.7 & 62.7 &  &  &  &  \\

\addlinespace[0.5ex] % Adjust the space as needed
\hdashline
\addlinespace[0.5ex] % Adjust the space as needed

\multirow{2}{*}{\shortstack{Dataflow-\\ Simple}} & No DD    &      &      &      &      & 50.9 & 22.7 &      &      \\
 % & Enumeration only                             &      &      &      &      & 63.0 & 50.0 &      &      \\
 & Full DD                                         &      &      &      &      & 59.7 & 63.9 &      &      \\
\midrule
\multirow{2}{*}{\shortstack{$\lambda$-DCS \\ Simple}} & No DD    &      &      &      &      &  &  &  37.3 & 27.5  \\
 & Full DD                                         &      &      &      &      &  &  &  38.0 & 31.7   \\
\bottomrule
\end{tabular}
\end{threeparttable}

% 61.8 & 55.1 & 57.5 & 41.8
% 65.5 & 62.3 & 70.1 & 59.3
% 65.7 & 56.6 & 61.8 & 45.6
% 73.1 & 68.7 & 75.7 & 62.7

\caption{Development set execution accuracies for Python, Javascript, Scala and SQL comprising 2.5\%, 19.6\%, 0.1\% and 4.9\% of Stack \citep{kocetkov2023the}, respectively, along with two DSL variations. There is no clear winner among the various PLs, suggesting that the prevalence in pretraining corpora is not a good predictor of performance.  $^\dagger$ For SQL we use the schema definition, see App.~\ref{app:dd}.
% along with the percentage of their prevalence in the Stack, a popular code corpus, along with
}
\label{tab:results-pls}
\end{table}

\subsection{What Makes a Good MR?}
\label{subsec:why-python}

Building on the findings from Section \ref{subsec:python-vs-dsl}, showing that Python prompts consistently outperform DSLs, we now investigate the source of these performance gains. Specifically, in this section, we ask:
\begin{enumerate}[itemsep=0pt,itemsep=4pt,parsep=0pt,topsep=2pt]
    \item Is the performance gain of a PL linked to its prevalence in pretraining corpora? 
    \item Can rare DSLs be simplified in a way that enables them to perform as well as PLs?
    \item Does the ability to break down programs into intermediate steps contribute to the improved performance of PLs?
\end{enumerate}

\subsubsection{Effect of a PL's Prevalence}
\label{subsec:other-pls}
To answer the first question, 
% we extend our experiments to include Scala, Javascript and SQL (SQL queries are only available for GeoQuery). 
we extend our experiments to include Scala and Javascript. For GeoQuery, which requires querying a database, we additionally experiment with SQL, a common query language.

According to the PL distribution provided by the Stack \cite{kocetkov2023the}, a large corpus of GitHub code, Scala is far less common than Python (0.1\% vs 2.5\%), while Javascript and SQL are more popular (19.6\% and 4.9\%).

\paragraph{Evaluation Procedure for Additional PLs.} We evaluate the performance of these additional PLs by prompting LLMs similarly to how we evaluated Python prompts (\S\ref{subsec:python-vs-dsl}). However, to avoid the undue engineering effort of implementing a complete executable environment for Scala and Javascript, as we did for Python, we evaluate generated programs by first automatically converting them to Python during inference time, similar to previous work \citep{cassano2023knowledge}, while confirming that the conversion is \textit{faithful} and does not introduce bias. For SQL, we use the original dataset queries\footnote{Taken from \url{https://github.com/jkkummerfeld/text2sql-data}} and use the schema definition instead of a Domain Description. We provide the complete procedure, prompts and analysis in App.~\ref{app:annotations:pls}.

% During evaluation, we convert predictions using GPT-4, with a prompt that includes both the Python and the target language's DDs and 15 conversion demonstrations. We test the correctness of converted Python programs by executing them within the Python environment. 
% To obtain a pool of demonstrations in the target language, we use a similar iterative procedure as the one described for Python in \Cref{subsec:experimental-setup}. 

\paragraph{Results.} Table~\ref{tab:results-pls} demonstrates that all three PLs outperform DSL-based prompts. However, the performance of the three PLs varies across datasets and splits, with Scala performing best in most splits, and SQL performing worst. This suggests that the prevalence of a PL in pretraining corpora alone does not reliably predict performance in semantic parsing tasks. This finding offers a subtle counterpoint to the results of \citet{cassano2023knowledge}, who identified a correlation between the prevalence of a PL in pretraining data and performance on other programming benchmarks.

\subsubsection{Simplifying PLs}
If the prevalence of PLs in pretraining corpora doesn't correlate with performance, could it be that DSL-based prompts perform worse because DSLs are overly complex, and simplifying them could improve performance \cite{herzig2021unlocking,li-etal-2022-exploring-secrets}? 

To investigate this, we experiment with simplified versions of SMCalFlow and Overnight's DSLs. Specifically, we use Dataflow-Simple \cite{meron2022simplifying}, a version of Dataflow tailored for creating events and querying organizational charts, which uses fewer operators and an entirely different syntax, with function calls in the style of popular PLs. While Dataflow-Simple isn't equivalent to Dataflow, it can be used to satisfy all of the requests in SMCalFlow's dataset. For Overnight, we create a simplified version of $\lambda$-DCS, where we remove redundant operators in the context of the evaluation setup, reducing its length by 42\% on average. Specifically, we remove the \texttt{call} operator, typing (\texttt{string}, \texttt{date}, \texttt{number}), redundant parentheses and the namespace \texttt{SW}. Examples for both MRs are provided in Table~\ref{tab:datasets}.

The results presented in the bottom two sections of Table~\ref{tab:results-pls} reveal that the surface-level simplification of $\lambda$-DCS provides only a marginal boost to performance. On the other hand, Dataflow-simple surprisingly performs nearly as well as the other PLs. These findings suggest that designing DSLs to resemble PLs could also be effective (when DD is included), even when DSLs are rare in pretraining corpora. However, what unique elements of PLs should be adopted in DSLs to yield comparable performance gains remains an open question.

\begin{table}[t]
    \centering
    \scriptsize
    
    \begin{threeparttable}
\setlength{\tabcolsep}{3.8pt}
\begin{tabular}{@{}lcccccccc@{}}
\toprule
 & \multicolumn{4}{c}{\textbf{GeoQuery}} & \multicolumn{2}{c}{\textbf{SMCalFlow-CS}} & \multicolumn{2}{c}{\textbf{Overnight}} \\
 \cmidrule[0.4pt](r{0.125em}){2-5}%
 \cmidrule[0.4pt](lr{0.125em}){6-7}%
 \cmidrule[0.4pt](lr{0.125em}){8-9}%
 &  i.i.d. & Templ. & TMCD & Len. & i.i.d. & 0-C & i.i.d. & Templ. \\ 
\midrule
Single & 83.2          & \textbf{80.9} & \textbf{80.3} & 79.1 & 64.8 & 60.4 & 62.9 & 57.6 \\
Multi.  & \textbf{83.7} & 80.6          & 80.0 & \textbf{80.3} & \textbf{67.7} & \textbf{70.0} & \textbf{64.5} & \textbf{72.9} \\
\bottomrule
\end{tabular}
\end{threeparttable}%

\caption{Accuracy of single-line programs against multiple-line programs with intermediate steps, in the Full DD setup. Breaking down code into intermediate steps usually contributes to performance, yet single line demonstrations still outperform DSL-based prompts.
}
\label{tab:results-decomposing}
\end{table}

\subsubsection{Effect of Intermediate Steps}
\label{subsec:single-vs-multi}
A key distinction between the PLs and the DSLs evaluated in this work lies in the fact that PLs allow breaking down the programs into multiple steps and assigning intermediate results to variables. To measure the impact of this aspect, we modify PL programs such that if a program contains more than one line, we compress it into a single line, eliminating intermediate variables and vice versa. We employ GPT-4 to perform these modifications and use execution-based evaluation to ensure that the program meaning does not change (see App.~\ref{app:annotations:singlue-multi} for the exact prompt). We note that the only modification made is to the programs of the prompt demonstrations, however, models can still output a program of any line length.

Results presented in Table~\ref{tab:results-decomposing} suggest that breaking down code into intermediate steps indeed contributes to higher performance in most cases. However, even single line demonstrations still significantly outperform DSL-based prompts.
\section{Conclusions}
In this work, we have shown that leveraging PLs and DDs does not only improve the effectiveness of in-context learning for semantic parsing, leading to substantial accuracy improvements across various datasets, but also significantly narrows the performance gap between i.i.d.\ and compositional splits and reduces the need for large demonstration pools. 
Our findings carry significant implications for the development of semantic parsing applications using modern LLMs. 
% More generally, the idea of using a PL for output representation and for specifying background information and task structure (as in DDs) is applicable beyond semantic parsing, and may be applicable to any generative tasks where the output must conform to some structure and has a step-wise nature (e.g., recipes, travel itineraries).

\section*{Limitations}

We evaluate models and methods using executable environments that we have implemented in Python; however, these implementations might not always accurately replicate the original environment. Particularly in SMCalFlow, which includes many long-tail operators infrequently used in the dataset, we omit some operators in our implementation.

We use OpenAI's API to annotate most of the Python programs that are used as demonstrations. While we validate the correctness of all programs, it is possible that this method introduces some bias into the nature of the generated programs.

We present the prevalence of different PLs in the Stack, assuming it offers a rough estimate of these languages' popularity on the web. However, the actual prevalence of PLs specifically within the training data of OpenAI's models, employed in this work, remains unknown.
Further, while our experiments with simplified versions of DSLs and rare PLs suggest that the improved performance of LLMs with PLs in compositional settings is not merely due to surface-level memorization, how much of these can be attributed to LLMs' ability to generalize compositionally versus memorization from pre-training corpus remains an important open question.

Finally, while this study focused on semantic parsing, the idea of using a PL for output representation and for specifying background information and task structure (as in DDs) could be applicable to any other generative tasks where the output must conform to some structure and has a step-wise nature (e.g., recipes, travel itineraries). We leave it to future work to explore these settings.

% \section*{Acknowledgements}

% We thank Alexander Koller and Matt Gardner for their helpful comments. This work was sponsored in part by the DARPA MCS program under Contract No.\ N660011924033 with the United States Office Of Naval Research and in part by the NSF award \#IIS-2046873. The views expressed are those of the authors and do not reflect the policy of the funding agencies.

% Entries for the entire Anthology, followed by custom entries
\bibliography{anthology,custom-rebiber}
\bibliographystyle{style/acl_natbib}

% \appendix
\clearpage
\appendix
\section{Datasets}

\subsection{Executable Environments}
\label{app:envs}

We describe the executable environments we use separately for each dataset and formalism.
\subsubsection{Geoquery}

% \lipsum[120]
\spacedparagraph{FunQL} To execute the FunQL queries, we use the GeoQuery\footnote{\url{https://www.cs.utexas.edu/users/ml/nldata/geoquery.html}} system, a prolog-based implementation that we execute using SWI-Prolog\footnote{\url{https://www.swi-prolog.org/}}. % The domain description for FunQL comprises a list of various functions exposed by the GeoQuery system and is shown in Figures \ref{fig:geoquery-funql}.

\paragraph{Python} We manually write a Python environment that is functionally equivalent to the GeoQuery system. The environment includes two components: a class for parsing and loading the Geobase database and an API for executing queries against this database. We show the API in Figures \ref{fig:geoquery-python} and \ref{fig:geoquery-python-2}. 

\paragraph{SQL} We use the SQLite engine to run SQL queries, with the data and schema provided in \url{https://github.com/jkkummerfeld/text2sql-data}.

\paragraph{Evaluation} Running queries with the GeoQuery system using FunQL, SQL and Python programs results in either a numeric result of a set of entities. We evaluate FunQL and Python programs by comparing their denotation against the gold denotation obtained by executing the gold FunQL program for each query, with no importance to order, and similarly evaluate SQL programs by comparing their denotation of gold SQL programs.

\subsubsection{SMCalFlow}

\spacedparagraph{Dataflow and Dataflow-Simple} We use the software provided by \citet{meron2022simplifying}\footnote{\url{https://github.com/telepathylabsai/OpenDF}} to execute Dataflow-Simple. Dataflow programs are executed by `simplifying' them, i.e. converting them to Dataflow-Simple, using the code provided in that package. The environment holds a database with people, the relationship between them in the organization, and a list of events.

\paragraph{Python}
We run Python programs by automatically converting them to Dataflow-Simple in a determinstic method, then executing them as mentioned above. Conversion is done by implementing each of the python classes and operators with a method that returns an AST that represents a relevant Dataflow-Simple sub-tree. For example, the Python method \texttt{find\_manager\_of(`person')} returns the corresponding AST of Dataflow-Simple's method, \texttt{FindManager(`person')}.

\paragraph{Evaluation}
All of the test instances in the splits we work with are requests to create events. Thus, to evaluate programs, we compare if the events created after running a generated program is exactly the same as the event create after running the gold Dataflow program. Since programs are executed using a database, which is used, for example, to find people by their names, we populate the database with a short list of people with random names. During evaluation, we extract names of people from both generated and gold programs, and arbitrarily map and replace each name in the programs to one of the people in the database. We do this for both generated and gold programs, while making sure that mapping is consistent in both of them during an evaluation for a single example.

We ignore the generated subject of the meeting, as we found that there are many inconsistencies in the way subjects were annotated: underspecified requests such as \textit{Set up a meeting with John} are often be annotated inconsistently, having either no subject, the subject ``meeting'', or something else.

\subsubsection{Overnight}

\spacedparagraph{$\lambda$-DCS and $\lambda$-DCS-Simple}
To execute $\lambda$-DCS programs, we use Sempre.\footnote{\url{https://github.com/percyliang/sempre}} Specifically, we use the executable Java program provided by \citet{herzig-berant-2018-decoupling}.\footnote{\url{https://github.com/jonathanherzig/zero-shot-semantic-parsing/blob/master/evaluator/evaluator.jar}}

\paragraph{Python}
To create the Python environment, we first use Sempre to output all entities in the `social-network' domain. We implement the python environment to be executed over these loaded entities.

\paragraph{Evaluation}
Running the programs returns a list of entities. For all formalisms, we consider accuracy to be correct iff the list of entities is exactly the same as the list of entities returned by running the gold $\lambda$-DCS program.

\subsection{Splits}
\label{app:dataset-splits}
For GeoQuery, we use the splits provided by \citet{shaw-etal-2021-compositional}, comprising the original i.i.d.\ split and the compositional generalization splits (Template, TMCD and length). 

For SMCalFlow, we use the i.i.d.\ and compositional splits proposed by \citet{yin-etal-2021-compositional}. These compositional splits evaluate predictions for queries that combine two domains: event creation and organizational chart. Specifically, we use the hardest "0-C" split, where the training set contains examples only from each of the domains separately, with no single example that combines both domains. For experiments with 5 or more demonstrations, we make sure there are at least two demonstrations from each of the domains.

For Overnight, we take the i.i.d.\ split and a compositional split (specifically {\small\texttt{template/split\_0}}, selected arbitrarily) from those published in \citet{bogin-etal-2022-unobserved}. 

We used the development sets for each of the datasets only to make sure predicted programs were executed as expected. For Overnight, where such a set was unavailable, we used 50 examples from the training set.

For GeoQuery, we use the entire tests sets (of size ranging from 279 to 331), while for SMCalFlow and Overnight, we sample 250 examples from the test sets.

We sample in-context demonstrations from the pool of training examples for which we have Python annotations. For GeoQuery, we have 824 such annotated programs, for SMCalFlow 128 and for Overnight 60.
\section{Additional Results}
\label{app:additional_results}

\begin{table*}[!ht]
    \centering
    \footnotesize
    
    \begin{threeparttable}
\begin{tabular}{@{}p{1cm}lcccccccc@{}}
\toprule
 & & \multicolumn{4}{c}{\textbf{GeoQuery}} & \multicolumn{2}{c}{\textbf{SMCalFlow-CS}} & \multicolumn{2}{c}{\textbf{Overnight}} \\
 \cmidrule[0.4pt](r{0.125em}){3-6}%
 \cmidrule[0.4pt](lr{0.125em}){7-8}%
 \cmidrule[0.4pt](lr{0.125em}){9-10}%
 & &  i.i.d. & Templ. & TMCD & Len. & i.i.d. & 0-C & i.i.d. & Templ. \\ 
\midrule
\multirow{3}{*}{DSL} & No DD & 24.4 & 19.0 & 28.0 & 14.7 & 23.3 & 0.7 & 18.1 & 7.7 \\
 & List of operators                   & 39.8 & 27.2 & 36.2 & 32.3 & 18.4 & 0.3 & 23.2 & 9.9 \\
 & Full DD                               & 46.9 & 45.2 & 45.2 & 38.9 & 22.7 & 2.0 & 22.4 & 12.0 \\
\midrule
\multirow{4}{*}{Python} & No DD & 56.1 & 42.2 & 50.2 & 32.4 & 22.8 & 5.1 & 51.9 & \textbf{39.9} \\
 & List of operators                      & \textbf{73.3} & \textbf{70.1} & 70.7 & 61.7 & 22.4 & 13.2 & 55.5 & 38.1 \\
 & DD w/o typing                        & 73.0 & 70.0 & 69.7 & 62.3 & 35.1 & 25.2 & 56.1 & 36.8 \\
 & Full DD                                  & 73.2 & 69.7 & \textbf{75.2} & \textbf{68.6} & \textbf{43.7} & \textbf{33.2} & \textbf{56.9} & 36.9 \\
\bottomrule
\end{tabular}
\end{threeparttable}%

\caption{Execution accuracy of Starcoder, comparing Python-based prompts with DSL-based prompts, across different DD variations, with 5 in-context demonstrations.
Similarly to ChatGPT (\Cref{tab:results-GPT}), Starcoder used with Python-based prompts with Full DD is consistently better than with DSL-based prompts.
Test sets results.
}
\label{tab:results-starcoder}
\end{table*}

\subsection{Starcoder}
\label{app:results-starcoder}

Main results for Starcoder are presented in \Cref{tab:results-starcoder}. With $k=5$ Starcoder's performance is generally lower than ChatGPT's, however the main trends remain the same: Python-based prompts with Full DD outperform DSL-based prompts in all cases, and Python-based prompts with Full DD outperform No DD in all cases but one.
\begin{table*}[!ht]
    \centering
    \footnotesize
    
    \begin{threeparttable}
\begin{tabular}{@{}p{1cm}lcccccccc@{}}
\toprule
 & & \multicolumn{4}{c}{\textbf{GeoQuery}} & \multicolumn{2}{c}{\textbf{SMCalFlow-CS}} & \multicolumn{2}{c}{\textbf{Overnight}} \\
 \cmidrule[0.4pt](r{0.125em}){3-6}%
 \cmidrule[0.4pt](lr{0.125em}){7-8}%
 \cmidrule[0.4pt](lr{0.125em}){9-10}%
 & &  i.i.d. & Templ. & TMCD & Len. & i.i.d. & 0-C & i.i.d. & Templ. \\ 
\midrule
\multirow{3}{*}{DSL} & No DD           & 4.3 & 11.4 & 1.4 & 4.6 & 6.9 & 8.8 & 3.8 & 2.0 \\
 & List of operators                   & 3.2 & 10.1 & 0.3 & 0.9 & 2.1 & 9.0 & 9.9 & 6.1 \\
 & Full DD (formal)                    & 6.1 & 8.7 & 5.5 & 4.7 & - & - & - & - \\
 & Full DD (NL)                        & 3.9 & 5.4 & 2.4 & 5.3 & 6.2 & 6.9 & 5.8 & 7.1 \\
\midrule
\multirow{4}{*}{Python} & No DD        & 13.0 & 11.6 & 5.6 & 3.8 & 4.5 & 8.7 & 2.8 & 0.7 \\
 & List of operators                   & 2.2 & 0.9 & 2.5 & 1.9 & 5.0 & 10.7 & 9.3 & 10.2 \\
 & DD w/o typing                       & 2.5 & 2.8 & 1.7 & 2.0 & 3.7 & 9.7 & 2.9 & 4.6 \\
 & Full DD                             & 0.9 & 1.2 & 0.5 & 1.5 & 3.4 & 10.9 & 1.1 & 4.5 \\
\bottomrule
\end{tabular}
\end{threeparttable}%
% 6.9 & 8.8
% 2.1 & 9.0
% 6.2 & 6.9
% 4.5 & 8.7
% 5.0 & 10.7
% 3.7 & 9.7
% 3.4 & 10.9

\caption{Standard deviations for ChatGPT's accuracy in \Cref{tab:results-GPT}.
}
\label{tab:results-GPT-std}
\end{table*}

\subsection{Standard Deviations}
\label{app:results-std}
All reported accuracy figures are average values obtained from three different seeds. The standard deviations corresponding to \Cref{tab:results-GPT} are detailed in \Cref{tab:results-GPT-std}.

\subsection{Prediction Examples}
\label{app:examples}
Examples for failed predictions are presented in \cref{tab:errors}, and for correct predictions in \cref{tab:successes}.

\begingroup
\lstset{
    style=mystyle,
    language=Python,
    aboveskip=-8pt,
    belowskip=-8pt
}
\begin{table*}[ht]
\centering
\small
\begin{tabular}{p{0.2\linewidth}p{0.5\linewidth}p{0.2\linewidth}}
\toprule
Input & Predicted Python Program & Error \\
\midrule
\textit{Put lunch with Jane 's boss in her office down for next Wednesday at noon .} &
\begin{lstlisting}
def answer():
    jane = api.find_person("Jane")
    boss = jane.find_manager_of()
    api.add_event(Event(subject="Lunch with Jane's Boss", attendees=[jane, boss], starts_at=[DateTimeValues.NextWeek, DateTimeClause.time_by_hm(hour=12, am_or_pm="pm")], location="Jane's Office"))
\end{lstlisting} &
% Conversion to dataflow failed due to incorrect API use: 
Execution error due to incorrect API use:
\texttt{NextWeek} is not a valid \texttt{DateTimeValue}.
\\
\midrule
% conversion error: exclude
\textit{i need to have all of jake 's team except jennifer at my staff meeting this friday} &
\begin{lstlisting}
def answer():
    jake = api.find_person("Jake")
    team = jake.find_team_of()
    team.remove(api.find_person("Jennifer"))
    api.add_event(Event(subject="Staff Meeting", attendees=team, starts_at=[DateTimeClause.get_next_dow(day_of_week="Friday")]))
\end{lstlisting} &
% Conversion to dataflow failed due to incorrect API use: 
Execution error due to incorrect API use:
The right way to exclude people is to use {\scriptsize \texttt{attendees\_to\_avoid}} argument of the \texttt{Event} class.
\\
\midrule
% Meaning error: Skip
\textit{I am going to be with Abby and her Skip tomorrow} &
\begin{lstlisting}
def answer():
    abby = api.find_person("Abby")
    skip = abby.find_manager_of()
    api.add_event(Event(attendees=[abby, skip], starts_at=[DateTimeValues.Tomorrow]))
\end{lstlisting} &
Incorrect execution as the model did not understand \textit{skip} means manager's manager.
\\
\midrule
% meaning error: "boss"
\textit{create drinks with boss on Tuesday} &
\begin{lstlisting}
def answer():
    boss = api.find_person("boss")
    api.add_event(Event(subject="Drinks with Boss", starts_at=[DateTimeClause.get_next_dow(day_of_week="Tuesday")], attendees=[boss]))
\end{lstlisting} &
Incorrect execution as the model did not use \texttt{find\_manager\_of} to find the manager.
\\
% \midrule
%     % execution error: after (env - not supported)
%     \textit{I 'd like to meet with Brit 's team after my email meeting .} &
% \begin{lstlisting}
% def answer():
%     brit = api.find_person("Brit")
%     email_meeting = api.find_event(subject="Email Meeting")
%     team = brit.find_team_of()
%     api.add_event(Event(subject="Meeting with Brit's Team", starts_at=email_meeting.ends_at, attendees=team))
% \end{lstlisting} &
% Execution Error as the Python environment did not support finding an exiting event.
% \\
\bottomrule
\end{tabular}
\caption{Examples of common errors made by GPT-Turbo-3.5 on the compositional 0-C split of SMCalFlow. Majority of errors were due to the LLM failing to understand the input utterance or incorrectly using the API.}
\label{tab:errors}
\end{table*}

\begin{table*}[h]
\centering
\small
\begin{tabular}{p{0.1\linewidth}p{0.2\linewidth}p{0.6\linewidth}}
\toprule
Dataset & Input & Predicted Python Program \\
\midrule
GeoQuery &
\textit{Which states border states through which the Mississippi traverses ?} &
\begin{lstlisting}
def answer() -> List[State]:
    mississippi_river = geo_model.find_river("mississippi")
    bordering_states = set()
    for state in mississippi_river.traverses:
        bordering_states.update(state.next_to)
    return list(bordering_states)
\end{lstlisting} \\
% \midrule
% SMCalFlow &
% \textit{Schedule lunch with Jennifer Watts and her team for tomorrow at noon .} &
% \begin{lstlisting}
% def answer():
%     jennifer = api.find_person("Jennifer Watts")
%     team = jennifer.find_team_of()
%     attendees = [jennifer] + team
%     api.add_event(Event(subject="Lunch", attendees=attendees, starts_at=[DateTimeValues.Tomorrow, DateTimeValues.Noon]))
% \end{lstlisting} \\
\midrule
SMCalFlow &
\textit{Schedule a meeting with my team on November 3rd at 11 am as well .} &
\begin{lstlisting}
def answer():
    team = api.get_current_user().find_team_of()
    api.add_event(Event(subject="Meeting with Team", starts_at=[DateTimeClause.date_by_mdy(month=11, day=3), DateTimeClause.time_by_hm(hour=11, am_or_pm="am")], attendees=team))
\end{lstlisting} \\
\midrule
Overnight &
\textit{student whose start date is end date of employee alice} &
\begin{lstlisting}
def answer():
    alice = api.find_person_by_id("en.person.alice")
    students_with_same_start_date = [person for person in api.people if person.education and any(e.start_date == alice_employment.end_date for e in person.education for alice_employment in alice.employment)]
    return students_with_same_start_date
\end{lstlisting} \\
\bottomrule
\end{tabular}
\caption{Examples of correct Python predictions made by GPT-Turbo-3.5 on the compositional TMCD split of GeoQuery, 0-C split of SMCalFlow, and Template split of Overnight. }
\label{tab:successes}
\end{table*}
\endgroup

\subsection{Exact Match Accuracy}
\label{subsec:exact_match_results}
We provide results for all DSL experiments with exact match as the metric for reference in Table~\ref{tab:results-GPT-exact-match}. Note that for Geoquery, while Full DD leads to significant improvements in execution accuracy (Table~\ref{tab:results-GPT}), when measuring exact match we see less of an improvement (e.g. 37.6 to 61.0 vs 20.7 to 27.6 in the i.i.d. split). We find that this is due to correct but different usage of the DSL, e.g. the model generates {\scriptsize \texttt{answer(count(traverse\_2(stateid('colorado'))))}}, which is different from the gold program {\scriptsize \texttt{answer(count(river(loc\_2(stateid('colorado')))))}}.

\begin{table*}[!ht]
    \centering
    \footnotesize
    
    \begin{threeparttable}
\begin{tabular}{@{}p{1cm}lcccccccc@{}}
\toprule
 & & \multicolumn{4}{c}{\textbf{GeoQuery}} & \multicolumn{2}{c}{\textbf{SMCalFlow-CS}} & \multicolumn{2}{c}{\textbf{Overnight}} \\
 \cmidrule[0.4pt](r{0.125em}){3-6}%
 \cmidrule[0.4pt](lr{0.125em}){7-8}%
 \cmidrule[0.4pt](lr{0.125em}){9-10}%
 & &  i.i.d. & Templ. & TMCD & Len. & i.i.d. & 0-C & i.i.d. & Templ. \\ 
\midrule
\multirow{3}{*}{DSL} & No DD         & 20.7 & 16.5 & 26.1 & 14.8 & 16.7 & 3.1 & 26.4 & 0.3 \\
 & List of operators                      & 28.7 & 13.5 & 29.4 & 18.3 & 17.3 & 4.3 & 29.2 & 0.5 \\
 & Full DD                            & 27.6 & 17.8 & 31.0 & 16.1 & 15.7 & 4.5 & 27.3 & 0.3 \\
\bottomrule
\end{tabular}
\end{threeparttable}%
\caption{Exact match accuracy of GPT-3.5-turbo for DSL-based prompts. Test set results.}
\label{tab:results-GPT-exact-match}
\end{table*}

\subsection{Accuracy vs \# of Tokens}
\label{app:accuracy_vs_tokens}
\begin{figure*}
  \centering
  \includegraphics[width=0.95\linewidth]{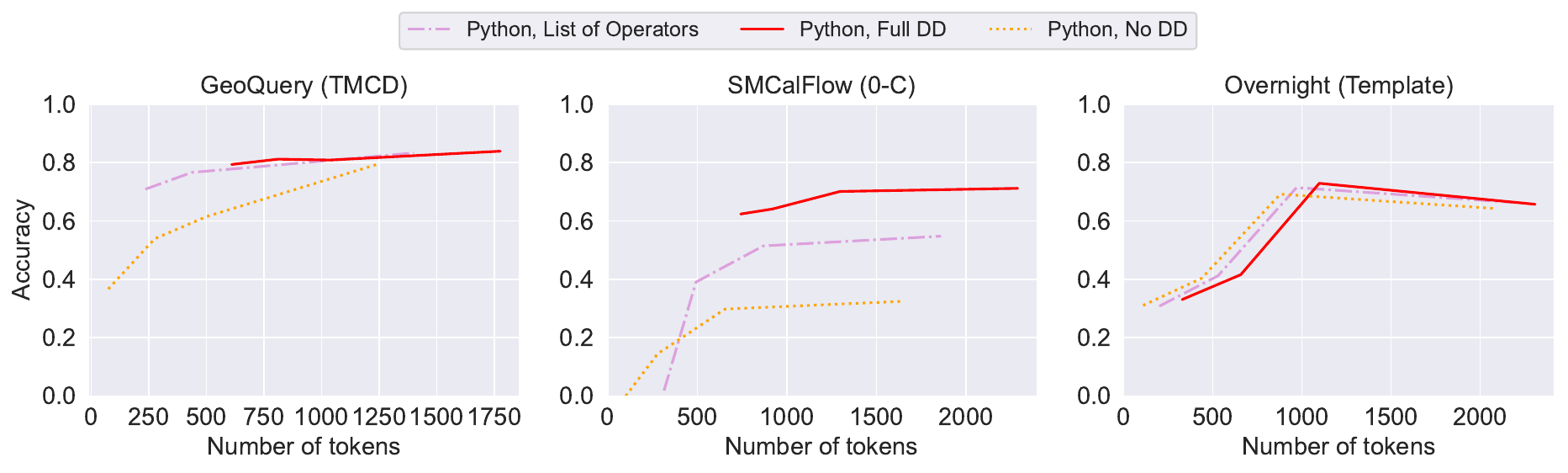}
  \caption{
  Execution accuracy for varying number of demonstrations, presenting the same data as Figure~\ref{fig:n-demonstrations-result} but visualizes it against the number of prompt tokens. The effect of DDs greatly varies between the datasets. For both GeoQuery and SMCalFlow, having the Full DD is preferred whenever it can fit.
  }
  \label{fig:n-tokens-result}
\end{figure*}

We present execution-based accuracy against the number of prompt tokens in Figure~\ref{fig:n-tokens-result} for three Python DD variations.

\section{Program Annotations}
\label{app:annotations}

\subsection{Python}
\label{app:annotations:python}

\begin{figure*}
    \centering
\begin{minted}{text}
Given the following data structures and functions:
[DD]

Write code to solve the following queries:

query: [query-1]
solution: [solution-1]
...
query: [query-test]
\end{minted}
    \caption{The prompt template we use. [DD] is replaced with the domain description for the environment being used, [query-$i$] and [solution-$i$] are replaced with utterance/output demonstrations, and [query-test] is replaced with the test utterance. Lines 1-3 are only included in experiments that contain DD.
    }
    \label{fig:prompt-template}
\end{figure*}

\begingroup
\begin{figure*}
    \centering
\begin{minted}{text}
Given the following python data structures and functions:

[Python DD]

and the corresponding javascript data structures and functions:

[Javascript DD]

convert the following javascript functions to python:

Javascript:
```javascript
[javascript-code-1]
```

Python:
```python
[python-code-1]
```
...

Javascript:
```javascript
[query-javascript-code]
```
Python:
```python
\end{minted}
    \caption{The prompt template we use to convert non-Python programs (Javascript in this case) to Python for evaluation. [Python DD] and [Javascript DD] are replaced with the corresponding domain descriptions, [javascript-code-$i$] and [python-code-$i$] with demonstrations of javascript to python conversion, and [query-javascript-code] is replaced with test Javascript code to be converted. }
    \label{fig:annotation-prompt-template}
\end{figure*}

\begin{figure*}
    \centering
\begin{minted}{text}
Given the following request and python program:

request: [query]

```python
[program]
```
We want to decompose the program into multiple lines as much as is reasonable possible.

```python
\end{minted}
    \caption{Prompt used to convert single-line programs to multiple-line programs.
    }
    \label{fig:prompt-singlemulti}
\end{figure*}

\begin{figure*}
    \centering
\begin{minted}{text}
Given the following request and python program:

request: [query]

```python
[program]
```
We want to make the python program a single line program that returns the same output.
If a single line program is not possible, use a minimal number of lines.

```python
\end{minted}
    \caption{Prompt used to convert multiple-line programs to single-line programs.
    }
    \label{fig:prompt-multisingle}
\end{figure*}

To create the pool of python programs for our experiment, we start by manually convert 2-10 examples to Python programs to seed our pool of Python-annotated instances. We then iteratively sample demonstrations from the pool and prompt an LLM with the Python DD (\S\ref{sec:python}) to automatically annotate the rest of the examples (we use either OpenAI's gpt-3.5-turbo or gpt-4\footref{footnote:chatgpt}). Only predictions that are evaluated to be correct, using the same execution-based evaluation described above, are added to the pool (see App.~\ref{app:annotations} for further details).

We use the prompt in Figure \ref{fig:prompt-template} with Python DD to generate Python programs.

\subsection{Scala and Javascript}
\label{app:annotations:pls}
We use the prompt in Figure \ref{fig:prompt-template} with the Scala or Javascript DD to generate programs for the corresponding language. To further convert to Python for execution-based evaluation, we use the prompt in Figure \ref{fig:annotation-prompt-template}. Tables \ref{tab:geoquery-js-conversion} and \ref{tab:geoquery-sc-conversion} contain example conversions from Javascript and Scala respectively to Python for GeoQuery.

To confirm that the conversion is \textit{faithful} and does not introduce bias, such as fixing incorrect programs or breaking correct ones, we manually analyzed 100 random examples of the converted Python programs, 50 each from Javascript and Scala, finding only 1 instance each of an unfaithful conversion.

\begingroup
\definecolor{codegreen}{rgb}{0,0.6,0}
\definecolor{codegray}{rgb}{0.5,0.5,0.5}
\definecolor{codepurple}{rgb}{0.58,0,0.82}
\definecolor{backcolour}{rgb}{0.95,0.95,0.92}
\lstset{
    style=mystyle,
    language=Python,
    aboveskip=-8pt,
    belowskip=-8pt
}
\begin{table*}[h]
\centering
\small
\begin{tabular}{p{0.1\linewidth}p{0.25\linewidth}p{0.28\linewidth}p{0.25\linewidth}}
\toprule
Input & Directly Annotated Python Program & Javascript Program & Converted Python Program \\
\midrule
\textit{In which state does the highest point in USA exist ?} &
\begin{lstlisting}
def answer() -> State:
    highest_point = max(geo_model.places, key=lambda x: x.elevation)
    return highest_point.state
\end{lstlisting} &
\begin{lstlisting}[language=Javascript]
function answer() {
    let highest_point_state = geo_model.states.reduce((maxState, currentState) => {
        return currentState.high_point > maxState.high_point ? currentState : maxState;
    });
    return highest_point_state.name;
}
\end{lstlisting} &
\begin{lstlisting}
def answer() -> str:
    highest_point_state = max(geo_model.states, key=lambda x: x.high_point.elevation)
    return highest_point_state.name
\end{lstlisting}
\\
\textit{What rivers flow through the largest state ?} &
\begin{lstlisting}
def answer() -> List[River]:
    largest_state = max(geo_model.states, key=lambda x: x.size)
    return largest_state.rivers
\end{lstlisting} &
\begin{lstlisting}[language=Javascript]
function answer() {
    let largest_state = geo_model.states.reduce((maxState, currentState) => {
        return currentState.area > maxState.area ? currentState : maxState;
    });
    return largest_state.rivers;
}
\end{lstlisting} &
\begin{lstlisting}
def answer() -> List[River]:
    largest_state = max(geo_model.states, key=lambda x: x.area)
    return largest_state.rivers
\end{lstlisting}
\\
\textit{What is the highest point in the state with the capital Des Moines ?} &
\begin{lstlisting}
def answer() -> Place:
    state = [s for s in geo_model.states if s.capital and s.capital.name == "des moines"][0]
    return state.high_point
\end{lstlisting} &
\begin{lstlisting}[language=Javascript]
function answer() {
    let des_moines_state = geo_model.find_city("des moines").state;
    return des_moines_state.high_point;
}
\end{lstlisting} &
\begin{lstlisting}
def answer() -> Place:
    des_moines_state = geo_model.find_city("des moines").state
    return des_moines_state.high_point
\end{lstlisting} \\
\bottomrule
\end{tabular}
\caption{Example conversions of predicted Javascript programs for GeoQuery to Python along with original (direct) Python annotation. Converted programs can differ from the directly annotated Python programs as the conversions were performed without the original query and hence tended to be faithful to given Javascript programs.}
\label{tab:geoquery-js-conversion}
\end{table*}

\begin{table*}[h]
\centering
\small
\begin{tabular}{p{0.1\linewidth}p{0.25\linewidth}p{0.28\linewidth}p{0.25\linewidth}}
\toprule
Input & Directly Annotated Python Program & Scala Program & Converted Python Program \\
\midrule
\textit{Which rivers run through states bordering New Mexico ?} &
\begin{lstlisting}
def answer() -> List[River]:
    new_mexico_state = geo_model.find_state("new mexico")
    bordering_states = new_mexico_state.next_to
    rivers = []
    for state in bordering_states:
        rivers.extend(state.rivers)
    return rivers
}\end{lstlisting} &
\begin{lstlisting}[language=Scala]
def answer(): Option[List[River]] = {
    val newMexicoState = geoModel.findState("new mexico")
    val borderingStates = newMexicoState.map(_.nextTo)
    val rivers = borderingStates.map(_.flatMap(_.rivers))
    rivers
}
\end{lstlisting} &
\begin{lstlisting}
def answer() -> List[River]:
    new_mexico_state = geo_model.find_state("new mexico")
    bordering_states = new_mexico_state.next_to
    rivers = [r for state in bordering_states for r in state.rivers]
    return rivers
\end{lstlisting}
\\
\textit{What is the most populated state bordering Oklahoma ?} &
\begin{lstlisting}
def answer() -> State:
    oklahoma_state = geo_model.find_state("oklahoma")
    return max(oklahoma_state.next_to, key=lambda x: x.population)
\end{lstlisting} &
\begin{lstlisting}[language=Scala]
def answer(): Option[State] = {
    val oklahomaState = geoModel.findState("oklahoma")
    val borderingStates = oklahomaState.map(_.nextTo)
    borderingStates.flatMap(_.maxByOption(_.population))
}
\end{lstlisting} &
\begin{lstlisting}
def answer() -> State:
    oklahoma_state = geo_model.find_state("oklahoma")
    bordering_states = oklahoma_state.next_to
    return max(bordering_states, key=lambda x: x.population)
\end{lstlisting}
\\
\textit{Which river goes through the most states ?} &
\begin{lstlisting}
def answer() -> River:
    return max(geo_model.rivers, key=lambda x: len(x.traverses))
\end{lstlisting} &
\begin{lstlisting}[language=Scala]
def answer() -> def answer(): Option[River] = {
    val riverWithMostStates = geoModel.rivers.maxBy(_.traverses.length)
    riverWithMostStates
}
\end{lstlisting} &
\begin{lstlisting}
def answer() -> River:
    return max(geo_model.rivers, key=lambda x: len(x.traverses))
\end{lstlisting} \\
\bottomrule
\end{tabular}
\caption{Example conversions of predicted Scala programs for GeoQuery to Python along with original (direct) Python annotation. Converted programs can differ from the directly annotated Python programs as the conversions were performed without the original query and hence tended to be faithful to given Scala programs.}
\label{tab:geoquery-sc-conversion}
\end{table*}
\endgroup

\subsection{Single/Multi Line Conversion}
\label{app:annotations:singlue-multi}

As described in \S\ref{subsec:single-vs-multi}, we convert single-line programs to multiple lines programs with intermediate steps and vice-versa, using GPT-4. We make sure conversions are correct by validating the execution-based accuracy of converted programs; if programs are invalid, we regenerate programs with a a temperature of 0.4 until a correct solution is found. We use GPT-4 with the prompts provided in Fig.~\ref{fig:prompt-singlemulti} and Fig.~\ref{fig:prompt-multisingle}.

The following is an example for a conversion of a multi-line program given the utterance \textit{``Which states have points higher than the highest point in Colorado?''}. The original annotation:
\begin{lstlisting}[style=mystyle,language=Python,xleftmargin=10pt]
def answer():
    colorado_state = geo_model.find_state("colorado")
    highest_point_in_colorado = colorado_state.high_point.elevation
    states_with_higher_points = [s for s in geo_model.states if s.high_point.elevation > highest_point_in_colorado]
    return states_with_higher_points
\end{lstlisting} 

The converted annotation:
\begin{lstlisting}[style=mystyle,language=Python,xleftmargin=10pt]
def answer():
    return [s for s in geo_model.states if s.high_point.elevation > geo_model.find_state("colorado").high_point.elevation]
\end{lstlisting} 
\section{Demonstration Selection Methods}
\label{app:demo_selection}

We experiment with two demonstration selection methods. 

\paragraph{Operator Coverage} This method selects a single fixed set of demonstrations with maximal coverage of operators that are used for every test input. For this, we use a slightly modified version of the greedy set coverage algorithm of \citet{gupta2023coveragebased}, shown in Algorithm \ref{alg:fixedcovalgo}. Here, the set of structures $\mathcal{S}$ is the set of all unigram operators in given formalism, and the measure of set-coverage is defined as $\mathtt{setcov}(\mathcal{S}, Z) = \sum_{s \in \mathcal{S}} \max_{z \in Z} \mathbbm{1}\left[s \in S_z\right]$ where $S_z$ is the set of operators in the candidate demonstration $z$.

\paragraph{BM25} We use BM25 to retrieve the most similar instances and use as demonstrations for each test input. We use the \texttt{rank\_bm25}\footnote{\url{https://github.com/dorianbrown/rank_bm25}} package's implementation of the Okapi variant of BM25.

\begin{algorithm}[H]
\small
\caption{Greedy Optimization of Set Coverage} % with Sets
\label{alg:fixedcovalgo}
\begin{algorithmic}[1]
    \Require Instance pool $\mathcal{T}$; Set of structures $\mathcal{S}$; desired number of demonstrations $k$; coverage scoring function $\mathtt{setcov}$
    \State $Z \gets \emptyset$\Comment{Selected Demonstrations}
    \State $Z_{curr} \gets \emptyset$\Comment{Current Set Cover}
    \State $\mathtt{curr\_cov} \gets -\inf$
    \While{$|Z| < k$}
        \State $z^*, \mathtt{next\_cov} = \arg\max\limits_{z \in \mathcal{T}-Z} \mathtt{setcov} \left(\mathcal{S}, Z_{curr} \cup z\right)$
        % \State $\mathtt{next\_cov} \gets \score\left(x_{\text {test }}, Z_{curr} \cup z^*\right)$
        \If{$\mathtt{next\_cov} > \mathtt{curr\_cov}$}
        \Comment{Pick $z^*$}
            \State $\mathtt{curr\_cov} \gets \mathtt{next\_cov}$
            \State $Z \gets Z \cup z^*$
            \State $Z_{curr} \gets Z_{curr} \cup z^*$
        \Else
        \Comment{Or start new cover}
            \State $Z_{curr} \gets \emptyset$, ~~$\mathtt{curr\_cov} \gets -\inf$
        \EndIf
    \EndWhile
    \State \textbf{return} $Z$
\end{algorithmic}
\end{algorithm}

\section{Domain Descriptions}
\label{app:dd}

We provide the domain descriptions that we use for each environment in the following figures:

\begin{itemize}
    \item Geoquery: Python (\ref{fig:geoquery-python}, \ref{fig:geoquery-python-2}), FunQL (NL: \ref{fig:geoquery-funql}, formal: \ref{fig:geoquery-funql-formal}), Javascript (\ref{fig:geoquery-js}, \ref{fig:geoquery-js-2}), Scala (\ref{fig:geoquery-sc}). For SQL, instead of domain descriptions, we use the schema definition taken verbatim from \url{https://raw.githubusercontent.com/jkkummerfeld/text2sql-data/master/data/geography-schema.csv}.
    \item SMCalFlow: Python (\ref{fig:smcalflow-python}), Dataflow (\ref{fig:smcalflow-dataflow}, \ref{fig:smcalflow-dataflow-2}), Dataflow-Simple (\ref{fig:smcalflow-dataflow-simple}, \ref{fig:smcalflow-dataflow-simple-2}), Javascript (\ref{fig:smcalflow-js}, \ref{fig:smcalflow-js-2}), Scala (\ref{fig:smcalflow-sc}).
    \item Overnight: Python (\ref{fig:overnight-python}) $\lambda$-DCS (\ref{fig:overnight-lambda-dcs}, \ref{fig:overnight-lambda-dcs-2}), $\lambda$-DCS Simple (\ref{fig:overnight-lambda-dcs-simple}), Javascript (\ref{fig:overnight-js}), Scala (\ref{fig:overnight-sc}).
\end{itemize}

% Geoquery using Python (\ref{fig:geoquery-python}, \ref{fig:geoquery-python-2}) and FunQL (\ref{fig:geoquery-funql}); SMCalFlow using Python (\ref{fig:smcalflow-python}), Dataflow (\ref{fig:smcalflow-dataflow}, \ref{fig:smcalflow-dataflow-2}) and Dataflow-Simple (\ref{fig:smcalflow-dataflow-simple}, \ref{fig:smcalflow-dataflow-simple-2}); Overnight using Python (\ref{fig:overnight-python}) $\lambda$-DCS (\ref{fig:overnight-lambda-dcs}, \ref{fig:overnight-lambda-dcs-2}) and $\lambda$-DCS Simple (\ref{fig:overnight-lambda-dcs-simple}).

\begingroup
\begin{figure*}
    \centering
\begin{minted}{text}
```python
@dataclass
class State:
    name: str
    abbreviation: str
    country: Country
    area: int
    size: int
    population: int
    density: float
    capital: Optional[City]
    high_point: Place
    low_point: Place
    next_to: List[State]
    cities: List[City]
    places: List[Place]
    mountains: List[Mountain]
    lakes: List[Lake]
    rivers: List[River]

@dataclass
class City:
    name: str
    state: State
    country: Country
    is_capital: bool
    population: int
    size: int
    is_major: bool
    density: float

@dataclass
class Country:
    name: str
    area: int
    population: int
    density: float
    high_point: Place
    low_point: Place
    cities: List[City]
    states: List[State]
    places: List[Place]
    mountains: List[Mountain]
    lakes: List[Lake]
    rivers: List[River]

@dataclass
class River:
    name: str
    traverses: List[State]
    length: int
    size: int
    is_major: bool

@dataclass
class Place:
    name: str
    state: State
    elevation: int
    size: int

@dataclass
class Mountain:
    name: str
    state: State
    elevation: int

@dataclass
class Lake:
    name: str
    area: int
    states: List[State]
\end{minted}
    \caption{Domain description for Geoquery, using Python. Continued in Fig.~\ref{fig:geoquery-python-2}.
    }
    \label{fig:geoquery-python}
\end{figure*}

\begin{figure*}
    \centering
\begin{minted}{text}
@dataclass
class GeoModel:
    countries: List[Country]
    states: List[State]
    cities: List[City]
    rivers: List[River]
    mountains: List[Mountain]
    lakes: List[Lake]
    places: List[Place]

    def find_country(self, name: str) -> Country:
        ...

    def find_state(self, name: str) -> State:
        ...

    def find_city(self, name: str, state_abbreviation: str = None) -> City:
        ...

    def find_river(self, name: str) -> River:
        ...

    def find_mountain(self, name: str) -> Mountain:
        ...

    def find_lake(self, name: str) -> Lake:
        ...

    def find_place(self, name: str) -> Place:
        ...

geo_model = GeoModel()
```
\end{minted}
    \caption{Domain description for Geoquery, using Python. Continued from Fig.~\ref{fig:geoquery-python}.
    }
    \label{fig:geoquery-python-2}
\end{figure*}

\begin{figure*}
    \centering
\begin{minted}{text}
```
cityid(CityName,StateAbbrev)  # given a city name and state, return the city id
countryid(CountryName)  # given a country name, return the country id
placeid(PlaceName)  # given a place (lakes, mountains, etc.) name, return the place id
riverid(RiverName)  # given a river name, return the river id
stateid(StateName)  # given a state name, return the state id

capital(all)  # return all cities that are capitals
city(all)  # return all cities
lake(all)  # return all lakes
mountain(all)  # return all mountains
place(all)  # return all places
river(all)  # return all rivers
state(all)  # return all states

capital(items)  # given a set of cities, return those that are capitals 
city(p)  # given a set of items, return those that are cities
lake(p)  # given a set of items, return those that are lakes
major(p)  # given a set of items, return those that are considered of major size
mountain(p)  # given a set of items, return those that are mountains
place(p)  # given a set of items, return those that are places
river(p)  # given a set of items, return those that are rivers
state(p)  # given a set of items, return those that are states
area_1(p)  # given a set of items, return their areas' sizes
capital_1(p)  # given a set of states, return their capitals
capital_2(p)  # given a set of cities, return their states
elevation_1(p)  # given a set of places, return their elevations
elevation_2(E)  # given a set of elevations, return the places with those elevations
high_point_1(p)  # given a set of items, return their highest points
high_point_2(p)  # given a set of places, return the items with those places as their highest points
higher_2(p)  # given a set of places, return the places that are higher than them
loc_1(p)  #  given a set of items, return where each item is located 
loc_2(p)  #  given a set of items, return the items located there
longer(p)  # given a set of rivers, return those that are longer than them
lower_2(p)  # given a set of places, return the places that are lower than them
len(p)  # given a set of rivers, return their lengths
next_to_1(p)  # given a set of states, return the states that are next to them
next_to_2(p)  # given a set of states, return the states that this state is next to
population_1(p)  # given a set of cities or states, return their populations
size(p)  # given a set of items, return their sizes (area for state, population for city, length for river)
traverse_1(p)  # given a set of rivers, return the states they traverse
traverse_2(p)  # given a set of states, return the rivers that traverse them

answer(p)  # return as answer (always needed)
largest(p)  # given a set of items, return the item with the largest size
largest_one(area_1(p))  # given a set of items, return the item with the largest area
largest_one(density_1(p))  # given a set of items, return the item with the largest density
largest_one(population_1(p))  # given a set of items, return the item with the largest population
smallest(p)  # given a set of items, return the item with the smallest size
smallest_one(area_1(p))  # given a set of items, return the item with the smallest area
smallest_one(density_1(p))  # given a set of items, return the item with the smallest density
smallest_one(population_1(p))  # given a set of items, return the item with the smallest population
highest(p)  # given a set of items, return the item that is highest
lowest(p)  # given a set of items, return the item that is lowest
longest(p)  # given a set of items, return the item that is longest
shortest(p)  # given a set of items, return the item that is shortest
count(p)  # given a set of items, return the number of items in the set
most(pD)  # given a set of items, return the item that appears most frequently in the set
fewest(pD)  # given a set of items, return the item that appears fewest times in the set

exclude(p1, p2)  # given a set of items, return the items that are in p1 but not in p2
intersect(p1, p2)  # given a set of items, return the items that are in both p1 and p2
```
\end{minted}
    \caption{Domain description for Geoquery, using FunQL (NL).
    }
    \label{fig:geoquery-funql}
\end{figure*}

\begin{figure*}
    \centering
\begin{minted}{text}
```
def cityid(CityName: str, StateAbbrev: str) -> City: ...
def countryid(CountryName: str) -> Country: ...
def placeid(PlaceName: str) -> Place: ...
def riverid(RiverName: str) -> River: ...
def stateid(StateName: str) -> State: ...
def capital(places: List[Place]) -> List[City]: ...
def city(places: List[Place]) -> List[City]: ...
def lake(places: List[Place]) -> List[Lake]: ...
def mountain(places: List[Place]) -> List[Mountain]: ...
def place(places: List[Place]) -> List[Place]: ...
def river(places: List[Place]) -> List[River]: ...
def state(places: List[Place]) -> List[State]: ...
def major(places: List[Place]) -> List[Place]: ...
def area_1(state: State | List[State]) -> List[float]: ...
def capital_1(state: State | List[State]) -> List[City]: ...
def capital_2(city: City | List[City]) -> List[State]: ...
def density_1(state: State | List[State]) -> List[float]: ...
def elevation_1(place: List[Place]) -> List[floats]: ...
def elevation_2(elevation: float) -> List[Place]: ...
def high_point_1(state: State | List[State]) -> List[Place]: ...
def high_point_2(place: Place) -> List[State]: ...
def higher_2(place: Place) -> List[Place]: ...
def loc_1(place: Place | List[Place]) -> List[State]: ...
def loc_2(state: State | List[State]) -> List[Place]: ...
def longer(river: River) -> List[River]: ...
def lower_2(place: Place) -> List[Place]: ...
def len(river: River | List[River]) -> List[float]: ...
def next_to_1(state: State | List[State]) -> List[State]: ...
def next_to_2(state: State | List[State]) -> List[State]: ...
def population_1(state: State | List[State]) -> List[float]: ...
def size(place: List[State] | List[City]) -> List[float]: ...
def traverse_1(river: River | List[River]) -> List[State]: ...
def traverse_2(state: State | List[State] | Country | List[Country]) -> List[River]: ...
def largest(place: List[Place]) -> List[Place]: ...
def largest_one(lst: List[Place]) -> Place: ...
def smallest(place: List[Place]) -> List[Place]: ...
def smallest_one(lst: List[Place]) -> Place: ...

def highest(place: List[Place]) -> List[Place]: ...
def lowest(place: List[Place]) -> List[Place]: ...
def longest(place: List[Place]) -> List[River]: ...
def shortest(place: List[Place]) -> List[River]: ...
def count(place: List[Place]) -> List[int]: ...
def most(place: List[Place]) -> Place: ...
def fewest(place: List[Place]) -> Place: ...

def exclude(lst1: List[Place], lst2: List[Place]) -> List[Place]: ...
def intersect(lst1: List[Place], lst2: List[Place]) -> List[Place]: ...
```
\end{minted}
    \caption{Domain description for Geoquery, using FunQL (formal).
    }
    \label{fig:geoquery-funql-formal}
\end{figure*}

\begin{figure*}
    \centering
\begin{minted}{text}
class State {
    constructor(name, abbreviation, country, area, population, density, capital, high_point, low_point, next_to, cities, places, mountains, lakes, rivers) {
        this.name = name;
        this.abbreviation = abbreviation;
        this.country = country;
        this.area = area;
        this.population = population;
        this.density = density;
        this.capital = capital;
        this.high_point = high_point;
        this.low_point = low_point;
        this.next_to = next_to;
        this.cities = cities;
        this.places = places;
        this.mountains = mountains;
        this.lakes = lakes;
        this.rivers = rivers;
    }
}

class City {
    constructor(name, state, country, is_capital, population, size, is_major) {
        this.name = name;
        this.state = state;
        this.country = country;
        this.is_capital = is_capital;
        this.population = population;
        this.size = size;
        this.is_major = is_major;
    }
}

class Country {
    constructor(name) {
        this.name = name;
    }
}

class River {
    constructor(name, traverses, length, size, is_major) {
        this.name = name;
        this.traverses = traverses;
        this.length = length;
        this.size = size;
        this.is_major = is_major;
    }
}

class Place {
    constructor(name, state, elevation, size) {
        this.name = name;
        this.state = state;
        this.elevation = elevation;
        this.size = size;
    }
}

class Mountain {
    constructor(name, state, elevation) {
        this.name = name;
        this.state = state;
        this.elevation = elevation;
    }
}

class Lake {
    constructor(name, area, states) {
        this.name = name;
        this.area = area;
        this.states = states;
    }
}
\end{minted}
    \caption{Domain description for Geoquery, using Javascript. Continued in Fig.~\ref{fig:geoquery-js-2}.
    }
    \label{fig:geoquery-js}
\end{figure*}

\begin{figure*}
    \centering
\begin{minted}{text}
class GeoModel {
    constructor(countries, states, cities, rivers, mountains, lakes, places) {
        this.countries = countries;
        this.states = states;
        this.cities = cities;
        this.rivers = rivers;
        this.mountains = mountains;
        this.lakes = lakes;
        this.places = places;
    }

    find_country(name) {
        // ...
    }

    find_state(name) {
        // ...
    }

    find_city(name, state_abbreviation = null) {
        // ...
    }

    find_river(name) {
        // ...
    }

    find_mountain(name) {
        // ...
    }

    find_lake(name) {
        // ...
    }

    find_place(name) {
        // ...
    }
}

let geo_model = new GeoModel();
```
\end{minted}
    \caption{Domain description for Geoquery, using Javascript. Continued from Fig.~\ref{fig:geoquery-js}.
    }
    \label{fig:geoquery-js-2}
\end{figure*}

\begin{figure*}
    \centering
\begin{minted}{text}
```scala
case class Country(name: String)

case class State(name: String, abbreviation: String, country: Country, area: Int, population: Int, density: Float, capital: Option[City], highPoint: Place, lowPoint: Place, nextTo: List[State], cities: List[City], places: List[Place], mountains: List[Mountain], lakes: List[Lake], rivers: List[River])

case class City(name: String, state: State, country: Country, isCapital: Boolean, population: Int, size: Int, isMajor: Boolean)

case class River(name: String, traverses: List[State], length: Int, size: Int, isMajor: Boolean)

case class Place(name: String, state: State, elevation: Int, size: Int)

case class Mountain(name: String, state: State, elevation: Int)

case class Lake(name: String, area: Int, states: List[State])

class GeoModel {
  var countries: List[Country] = List()
  var states: List[State] = List()
  var cities: List[City] = List()
  var rivers: List[River] = List()
  var mountains: List[Mountain] = List()
  var lakes: List[Lake] = List()
  var places: List[Place] = List()

  def findCountry(name: String): Option[Country] = ???

  def findState(name: String): Option[State] = ???

  def findCity(name: String, stateAbbreviation: Option[String] = None): Option[City] = ???

  def findRiver(name: String): Option[River] = ???

  def findMountain(name: String): Option[Mountain] = mountains.find(_.name == name)

  def findLake(name: String): Option[Lake] = lakes.find(_.name == name)

  def findPlace(name: String): Option[Place] = places.find(_.name == name)
}

val geoModel = new GeoModel()
```
\end{minted}
    \caption{Domain description for Geoquery, using Scala. 
    }
    \label{fig:geoquery-sc}
\end{figure*}

\begin{figure*}
    \centering
\begin{minted}{text}
```python
@dataclass
class Person:
    name: str
    
    def find_team_of() -> List["Person"]:
        ...
    
    def find_reports_of() -> List["Person"]:
        ...
    
    def find_manager_of() -> "Person":
        ...

@dataclass
class Event:
    attendees: List[Person] = None
    attendees_to_avoid: List[Person] = None
    subject: Optional[str] = None
    location: Optional[str] = None
    starts_at: Optional[List[DateTimeClause]] = None
    ends_at: Optional[List[DateTimeClause]] = None
    duration: Optional["TimeUnit"] = None
    show_as_status: Optional["ShowAsStatus"] = None

DateTimeValues = Enum("DateTimeValues", ["Afternoon", "Breakfast", "Brunch", "Dinner", "Early", "EndOfWorkDay", "Evening",
    "FullMonthofMonth", "FullYearofYear", "LastWeekNew", "Late", "LateAfternoon", "LateMorning", "Lunch", "Morning",
    "NextMonth", "NextWeekend", "NextWeekList", "NextYear", "Night", "Noon", "Now", "SeasonFall", "SeasonSpring",
    "SeasonSummer", "SeasonWinter", "ThisWeek", "ThisWeekend", "Today", "Tomorrow", "Yesterday"])

class DateTimeClause:
    def get_by_value(date_time_value: DateTimeValues) -> "DateTimeClause": ...
    def get_next_dow(day_of_week: str) -> "DateTimeClause": ...
    def date_by_mdy(month: int = None, day: int = None, year: int = None) -> "DateTimeClause": ...
    def time_by_hm(hour: int = None, minute: int = None, am_or_pm: str = None) -> "DateTimeClause": ...
    def on_date_before_date_time(date: "DateTimeClause", time: "DateTimeClause") -> "DateTimeClause": ...
    def on_date_after_date_time(date: "DateTimeClause", time: "DateTimeClause") -> "DateTimeClause": ...
    def around_date_time(date_time: "DateTimeClause") -> "DateTimeClause": ...


TimeUnits = Enum("TimeUnits", ["Hours", "Minutes", "Days"])
TimeUnitsModifiers = Enum("TimeUnitsModifiers", ["Acouple", "Afew"])

@dataclass
class TimeUnit:
    number: Optional[Union[int,float]] = None
    unit: Optional[TimeUnits] = None
    modifier: Optional[TimeUnitsModifiers] = None

ShowAsStatusType = Enum("ShowAsStatusType", ["Busy", "OutOfOffice"])
    

class API:
    def find_person(name: str) -> Person:
        ...
    
    def get_current_user() -> Person:
        ...

    def add_event(event: Event) -> None:
        ...

    def find_event(attendees: Optional[List[Person]] = None, subject: Optional[str] = None) -> Event:
        ...

api = API()
```
\end{minted}
    \caption{Domain description for SMCalFlow, using Python.
    }
    \label{fig:smcalflow-python}
\end{figure*}

\begin{figure*}
    \centering
\begin{minted}{text}
```
Yield  # Arguments: (1) :output, the function to be executed. Returns: The result of the function.
CreateCommitEventWrapper   # Arguments: (1) :event, containing event details. Returns: The created event.
CreatePreflightEventWrapper  # Arguments: (1) :constraint, containing event details. Returns: The event that satisfies the constraint.
FindEventWrapperWithDefaults  # Arguments: (1) :constraint, the constraint to be satisfied by the event. Returns: The event that satisfies the constraint.

extensionConstraint  # Arguments: (1) the type of constraint (e.g., Constraint[Recipient], Constraint[Date], RecipientWithNameLike), Returns: A constraint that needs to be satisfied by the entity.
Constraint[Event]  # Arguments: (1) :attendees, :start, :subject or :location. Returns: Constraints to create or find an event.
Constraint[DateTime]  # Arguments: (1) :date, the date constraint. Returns: A constraint that needs to be satisfied by the date and time.
andConstraint  # Arguments: Any number of constraints. Returns: A constraint that is satisfied when all the input constraints are satisfied.
RecipientWithNameLike  # Arguments: (1) :constraint, the type of constraint (e.g., Constraint[Recipient]), (2) :name, the name of the recipient. Returns: A constraint that needs to be satisfied by the recipient.
PersonName  # Arguments: (1) the name of the person. Returns: The name of the person. e.g. `PersonName " Dan "`  
AttendeeListHasRecipient  # Arguments: (1) :recipient, the recipient to be included. Returns: A constraint for the event.
AttendeeListHasPeople  # Arguments: (1) :people, the group of people to be included. Returns: A constraint for the event.
AttendeeListHasRecipientConstraint  # Arguments: (1) :recipientConstraint, the recipient constrained to be included. Returns: A constraint for the event.
DateTimeConstraint  # Arguments: (1) :constraint, the time constraint, (2) :date, the date. Returns: A constraint that needs to be satisfied by the date and time.
AttendeeListExcludesRecipient  # Arguments: (1) :recipient, the recipient to be excluded. Returns: A constraint for the event.

Execute  # Arguments: (1) :intension, the intension to be executed. Returns: The entity referred to by the intension.
refer  # Arguments: (1) extensionConstraint, the constraint to be satisfied by the entity. Returns: A reference to the entity that satisfies the constraint.
singleton  # Arguments: (1) an element or a list with an element. Returns: The single element.
do  # Arguments: Any number of functions. Returns: The results of the functions.
String  # Arguments: (1) a literal string. Returns: A string representation.

FindManager  # Arguments: (1) :recipient, the recipient whose manager is to be found. Returns: The manager of the recipient.
FindReports  # Arguments: (1) :recipient, the recipient whose reports are to be found. Returns: The reports of the recipient.
FindTeamOf  # Arguments: (1) :recipient, the recipient whose team is to be found. Returns: The group of people who make up the recipient's team.
toRecipient  # Arguments: (1) A user. Returns: The given user as a recipient.
CurrentUser  # Arguments: None. Returns: The current user.

LocationKeyphrase  # Arguments: (1) the location. Returns: The location. e.g. `LocationKeyphrase " office "`  
roomRequest  # Arguments: None. Returns: A request for a room.

# These operators represent specific times or dates. They have no arguments and return the specified time or date.
Today
Tomorrow
NextWeekList
NextDOW
Noon
Afternoon
Morning
Night
EndOfWorkDay
Evening
Weekend
ThisWeekend
ThisWeek
Early
Now
NextYear
Lunch

# These operators represent specific numbers or convert values to numbers. Arguments: (1) the number or value to be converted. Returns: The specific number or the converted value.
Number
NumberAM
NumberPM
\end{minted}
    \caption{Domain description for SMCalFlow, using DataFlow. Continued in Fig.~\ref{fig:smcalflow-dataflow-2}.
    }
    \label{fig:smcalflow-dataflow}
\end{figure*}

\begin{figure*}
    \centering
\begin{minted}{text}
toDays
toHours
toMinutes
DateAndConstraint  # Arguments: (1) :date1, the first date, (2) :date2, the second date. Returns: The date and constraint.
DateAtTimeWithDefaults  # Arguments: (1) :date, the date, (2) :time, the time. Returns: The date and time.
nextDayOfMonth  # Arguments: (1) the day of the month. Returns: The next occurrence of the day of the month.
DayOfWeek  # Arguments: (1) the day of the week. Returns: The day of the week.
DowOfWeekNew  # Arguments: (1) :dow, the day of the week, (2) :week, the week. Returns: The day of the week in the week.
previousDayOfWeek  # Arguments: (1) :dayOfWeek, the day of the week. Returns: The previous occurrence of the day of the week.
NextDOW  # Arguments: (1) :dow, the day of the week. Returns: The next occurrence of the day of the week.
EventAllDayStartingDateForPeriod  # Arguments: (1) :event, the event, (2) :period, the duration of the event, (3) :startDate, the start date of the event. Returns: The event with the specified start date and duration.
PeriodDuration  # Arguments: (1) :duration, the duration. Returns: The period duration.

MD  # Arguments: (1) :day, the day, (2) :month, the month. Returns: The date.
MDY  # Arguments: (1) :day, the day, (2) :month, the month, (3) :year, the year. Returns: The date.
Month  # Arguments: (1) the month. Returns: The month.
NextTime  # Arguments: (1) :time, the time. Returns: The next occurrence of the time.
HourMinuteAm  # Arguments: (1) :hours, the hours, (2) :minutes, the minutes. Returns: The time.
HourMinutePm  # Arguments: (1) :hours, the hours, (2) :minutes, the minutes. Returns: The time.
FullMonthofMonth  # Arguments: (1) :month, the month. Returns: The full month.

TimeAfterDateTime  # Arguments: (1) :dateTime, the date and time, (2) :time, the time after the date and time. Returns: The time after the date and time.
OnDateAfterTime  # Arguments: (1) :date, the date, (2) :time, the time after the date. Returns: The date after the time.
AroundDateTime  # Arguments: (1) :dateTime, the date and time. Returns: The time around the date and time.
```
\end{minted}
    \caption{Domain description for SMCalFlow, using DataFlow. Continued from Fig.~\ref{fig:smcalflow-dataflow}.
    }
    \label{fig:smcalflow-dataflow-2}
\end{figure*}

\begin{figure*}
    \centering
\begin{minted}{text}
```
FindTeamOf  # given a person name or id, returns a pseudo-person representing the team of that person
FindReports  # given a person name or id, returns a pseudo-person representing the reports of that person
FindManager  # given a person name or id, returns the manager of that person

with_attendee  # given a person name or id, returns a clause to match or create an event with that person as an attendee
avoid_attendee  # given a person name or id, returns an event clause to avoid that attendee when creating an event
has_subject  # given a string, returns an event to match or create an event with that subject
at_location  # given a string, returns an event clause to match or create an event at that location
starts_at  # given a datetime clause, returns an event clause to match or create an event starting at that time
ends_at  # given a datetime clause, returns an event clause to match or create an event ending at that time
has_duration  # given a time unit value, returns an event clause to match or create an event with that duration
has_status  # given a ShowAsStatus value, returns an event clause to match or create an event with that status

# the following operators return datetime clauses and accept no arguments
Afternoon
Breakfast
Brunch
Dinner
Early
EndOfWorkDay
Evening
FullMonthofMonth
FullYearofYear
LastWeekNew
Late
LateAfternoon
LateMorning
Lunch
Morning
NextMonth
NextWeekend
NextWeekList
NextYear
Night
Noon
Now
SeasonFall
SeasonSpring
SeasonSummer
SeasonWinter
ThisWeek
ThisWeekend
Today
Tomorrow
Yesterday

# general date time clauses
DateTime  # given either a datetime clause representing a date and/or a time operator representing a time, returns a datetime clause
Date  # given a date or dayofweek, returns a date
DayOfWeek  # given a day of week string, returns a time clause
NextDOW  # given a day of week string, returns a time clause for the next occurrence of that day of week
MD  # given a month and day as arguments, returns a date clause
MDY  # given a month, day, and year as arguments, returns a date clause

# given a value, the following operators return datetime clauses according to the given value
toMonth
toFourDigitYear
HourMinuteAm
HourMinutePm
NumberAM
NumberPM

# given a datetime clause, the following operators modify the clause and return a datetime clause according to the modification
OnDateAfterTime
OnDateBeforeTime
AroundDateTime
\end{minted}
    \caption{Domain description for SMCalFlow, using DataFlow Simple. Continued in Fig.~\ref{fig:smcalflow-dataflow-simple-2}.
    }
    \label{fig:smcalflow-dataflow-simple}
\end{figure*}

\begin{figure*}
    \centering
\begin{minted}{text}
# given either a number or the operators Acouple/Afew, all the following operators return time unit values according to the given unit
toDays
toHours
toMinutes

# these operators can be used to create time unit values instead of using integer values
Acouple
Afew

ShowAsStatus  # enumeration of possible event statuses (Busy, OutOfOffice)

CreateEvent  # given multiple event clauses (such as with_attendee, has_subject, combined together with `AND`), creates an event complying with those clauses
FindEvents  # given multiple event clauses (such as with_attendee, has_subject, combined together with `AND`), returns a list of events complying with those clauses
CurrentUser  # returns the current user (person)

do  # allows the execution of multiple commands in a single prompt (each command is an argument). Often used in conjunction with `Let` to define variables
Let  # defines a variable (first argument) with a value (second argument)

AND  # combines multiple event clauses together
```
\end{minted}
    \caption{Domain description for SMCalFlow, using DataFlow Simple. Continued from Fig.~\ref{fig:smcalflow-dataflow-simple}.
    }
    \label{fig:smcalflow-dataflow-simple-2}
\end{figure*}

\begin{figure*}
    \centering
\begin{minted}{text}
```javascript
class Person {
    constructor(name) {
        this.name = name;
    }

    find_team_of() {
        // ...
    }

    find_reports_of() {
        // ...
    }

    find_manager_of() {
        // ...
    }
}

class Event {
    constructor(attendees = null, attendees_to_avoid = null, subject = null, location = null, starts_at = null, ends_at = null, duration = null, show_as_status = null) {
        this.attendees = attendees;
        this.attendees_to_avoid = attendees_to_avoid;
        this.subject = subject;
        this.location = location;
        this.starts_at = starts_at;
        this.ends_at = ends_at;
        this.duration = duration;
        this.show_as_status = show_as_status;
    }
}

const DateTimeValues = ["Afternoon", "Breakfast", "Brunch", "Dinner", "Early", "EndOfWorkDay", "Evening",
    "FullMonthofMonth", "FullYearofYear", "LastWeekNew", "Late", "LateAfternoon", "LateMorning", "Lunch", "Morning",
    "NextMonth", "NextWeekend", "NextWeekList", "NextYear", "Night", "Noon", "Now", "SeasonFall", "SeasonSpring",
    "SeasonSummer", "SeasonWinter", "ThisWeek", "ThisWeekend", "Today", "Tomorrow", "Yesterday"];

class DateTimeClause {
    get_by_value(date_time_value) {
        // ...
    }

    get_next_dow(day_of_week) {
        // ...
    }

    date_by_mdy(month = null, day = null, year = null) {
        // ...
    }

    time_by_hm(hour = null, minute = null, am_or_pm = null) {
        // ...
    }

    on_date_before_date_time(date, time) {
        // ...
    }

    on_date_after_date_time(date, time) {
        // ...
    }

    around_date_time(date_time) {
        // ...
    }
}

const TimeUnits = ["Hours", "Minutes", "Days"];
const TimeUnitsModifiers = ["Acouple", "Afew"];

\end{minted}
    \caption{Domain description for SMCalFlow, using Javascript. Continued in Fig.~\ref{fig:smcalflow-js-2}.
    }
    \label{fig:smcalflow-js}
\end{figure*}

\begin{figure*}
    \centering
\begin{minted}{text}
class TimeUnit {
    constructor(number = null, unit = null, modifier = null) {
        this.number = number;
        this.unit = unit;
        this.modifier = modifier;
    }
}

const ShowAsStatusType = ["Busy", "OutOfOffice"];

class API {
    find_person(name) {
        // ...
    }

    get_current_user() {
        // ...
    }

    add_event(event) {
        // ...
    }

    find_event(attendees = null, subject = null) {
        // ...
    }
}

const api = new API();
```
\end{minted}
    \caption{Domain description for SMCalFlow, using Javascript. Continued from Fig.~\ref{fig:smcalflow-js}.
    }
    \label{fig:smcalflow-js-2}
\end{figure*}

\begin{figure*}
    \centering
\begin{minted}{text}
```scala
case class Person(name: String) {
  def findTeamOf(): List[Person] = ???
  def findReportsOf(): List[Person] = ???
  def findManagerOf(): Person = ???
}

case class Event(var attendees: Option[List[Person]] = None,
                 var attendeesToAvoid: Option[List[Person]] = None,
                 var subject: Option[String] = None,
                 var location: Option[String] = None,
                 var startsAt: Option[List[DateTimeClause]] = None,
                 var endsAt: Option[List[DateTimeClause]] = None,
                 var duration: Option[TimeUnit] = None,
                 var showAsStatus: Option[ShowAsStatusType.Value] = None)

object DateTimeValues extends Enumeration {
  val Afternoon, Breakfast, Brunch, Dinner, Early, EndOfWorkDay, Evening,
  FullMonthofMonth, FullYearofYear, LastWeekNew, Late, LateAfternoon, LateMorning, Lunch, Morning,
  NextMonth, NextWeekend, NextWeekList, NextYear, Night, Noon, Now, SeasonFall, SeasonSpring,
  SeasonSummer, SeasonWinter, ThisWeek, ThisWeekend, Today, Tomorrow, Yesterday = Value
}

class DateTimeClause {
  def getByValue(dateTimeValue: DateTimeValues.Value): DateTimeClause = ???
  def getNextDow(dayOfWeek: String): DateTimeClause = ???
  def dateByMdy(month: Option[Int] = None, day: Option[Int] = None, year: Option[Int] = None): DateTimeClause = ???
  def timeByHm(hour: Option[Int] = None, minute: Option[Int] = None, amOrPm: Option[String] = None): DateTimeClause = ???
  def onDateBeforeDateTime(date: DateTimeClause, time: DateTimeClause): DateTimeClause = ???
  def onDateAfterDateTime(date: DateTimeClause, time: DateTimeClause): DateTimeClause = ???
  def aroundDateTime(dateTime: DateTimeClause): DateTimeClause = ???
}

object TimeUnits extends Enumeration {
  val Hours, Minutes, Days = Value
}

object TimeUnitsModifiers extends Enumeration {
  val Acouple, Afew = Value
}

case class TimeUnit(var number: Option[Either[Int, Double]] = None,
                    var unit: Option[TimeUnits.Value] = None,
                    var modifier: Option[TimeUnitsModifiers.Value] = None)

object ShowAsStatusType extends Enumeration {
  val Busy, OutOfOffice = Value
}

class API {
  def findPerson(name: String): Person = ???
  def getCurrentUser(): Person = ???
  def addEvent(event: Event): Unit = ???
  def findEvent(attendees: Option[List[Person]] = None, subject: Option[String] = None): Event = ???
}

val api = new API
```
\end{minted}
    \caption{Domain description for SMCalFlow, using Scala. 
    }
    \label{fig:smcalflow-sc}
\end{figure*}

\begin{figure*}
    \centering
\begin{minted}{text}
```python
Gender = Enum('Gender', 'male,female')
RelationshipStatus = Enum('RelationshipStatus', 'single,married')
Education = NamedTuple('Education', [('university', str), ('field_of_study', str), ('start_date', int), ('end_date', int)])
Employment = NamedTuple('Employment', [('employer', str), ('job_title', str), ('start_date', int), ('end_date', int)])

@dataclass
class Person:
    name: str
    gender: Gender
    relationship_status: RelationshipStatus
    height: int
    birthdate: int
    birthplace: str
    friends: List['Person'] = None
    logged_in: bool = False

    education: List[Education] = None
    employment: List[Employment] = None


@dataclass
class API:
    people: List[Block]
    
    def find_person_by_id(self, block_id: str) -> Person:
        ...

api = API()
```
\end{minted}
    \caption{Domain description for Overnight, using Python.
    }
    \label{fig:overnight-python}
\end{figure*}

\begin{figure*}
    \centering
\begin{minted}{text}
```
call  # invoke a function. Arguments: (1) function to be invoked, (2 and subsequent) parameters to be passed to that function or method. Returns: the result of the function call.
SW.listValue  # extract values from an object. Arguments: (1) An object of any type. Returns: A list of values.
SW.filter  # applies a filter to a list of objects. Arguments: (1) A list of objects, (2) A property to filter on, (3) A comparison operator, (4) A value to compare against. If property is boolean (unary), arguments: (1) A list of objects, (2) Unary property to filter on. Returns: A list of objects that pass the filter.
SW.getProperty  # retrieves a property from an object. Arguments: (1) An object, (2) A property name. Returns: The value of the property.
SW.reverse  # reverses the direction of a property. Arguments: (1) A property name. Returns: The reversed property name.
SW.singleton  # creates a singleton set containing a single object. Arguments: (1) An object. Returns: A singleton set containing the object.
SW.domain  # retrieves the domain of a property, which is the set of entities or objects that the property can be applied to. Arguments: (1) A property name. Returns: The set of entities that can have the property.
SW.countSuperlative  # finds the object(s) with the minimum or maximum count of a certain property. Arguments: (1) A list of objects, (2) A superlative operator (min or max), (3) A property to count, (4) A list of objects to count from. Returns: The object(s) with the minimum or maximum count of the property.
SW.ensureNumericProperty  # ensures that a property is treated as numeric for comparison purposes. Arguments: (1) A property name. Returns: The property name, treated as numeric.
SW.ensureNumericEntity  # ensures that an entity is treated as numeric for comparison purposes. Arguments: (1) An entity. Returns: The entity, treated as numeric.
SW.size  # retrieves the size of a collection. Arguments: (1) A collection of objects. Returns: The size of the collection as a numeric value.
SW.aggregate  # applies an aggregate function to a property over a set of objects. Arguments: (1) An aggregate function (e.g., sum, avg, min, max), (2) A property to aggregate over, (3) A set of objects. Returns: The result of the aggregate function.
SW.concat  # concatenates two or more strings or lists. Arguments: (1 and subsequent) Strings or lists to concatenate. Returns: The concatenated result.
SW.countComparative  # compares the count of a property over a set of objects with a given number. Arguments: (1) A set of objects, (2) A property to count, (3) A comparison operator, (4) A number to compare against, (5) A set of objects to count from. Returns: The objects for which the count of the property satisfies the comparison.
SW.superlative  # finds the object(s) with the minimum or maximum value of a certain property. Arguments: (1) A set of objects, (2) A superlative operator (min or max), (3) A property to compare. Returns: The object(s) with the minimum or maximum value of the property.

lambda  # creates a function. Arguments: (1) A variable name, (2) A function body. Returns: A function.
var  # references a variable. Arguments: (1) A variable name. Returns: The value of the variable.
string  # creates a string. Arguments: (1) A string value. Returns: The string.
number  # creates a number. Arguments: (1) A numeric value, (2) A unit (optional). Returns: The number.
date  # creates a date. Arguments: (1) Year, (2) Month, (3) Day. Returns: The date.

# The following are namespaces for different types of entities.
en.person
en.company
en.university
en.relationship_status
en.employee
en.student
en.field
en.city
en.gender

# specific entities under these namespaces:
en.gender.female
en.gender.male
en.relationship_status
en.relationship_status.single
en.relationship_status.married

# en.person properties:
height  # property of type (number with unit en.cm)
birthdate  # property of type date
birthplace  # property of type en.city
logged_in  # property of type bool
friend  # property of type en.person
relationship_status  # property of type en.relationship_status
```
\end{minted}
    \caption{Domain description for Overnight, using $\lambda$-DCS. Continued in Fig.~\ref{fig:overnight-lambda-dcs-2}.
    }
    \label{fig:overnight-lambda-dcs}
\end{figure*}

\begin{figure*}
    \centering
\begin{minted}{text}
# education properties:
student  # property of type en.person
university  # property of type en.university
field_of_study  # property of type en.field
education_start_date  # property of type date
education_end_date  # property of type date

# employment properties:
employee  # property of type en.person
employer  # property of type en.company
job_title  # property of type string
employment_start_date  # property of type date
employment_end_date  # property of type date
\end{minted}
    \caption{Domain description for Overnight, using $\lambda$-DCS. Continued from Fig.~\ref{fig:overnight-lambda-dcs}.
    }
    \label{fig:overnight-lambda-dcs-2}
\end{figure*}

\begin{figure*}
    \centering
\begin{minted}{text}
listValue  # extract values from an object. Arguments: (1) An object of any type. Returns: A list of values.
filter  # applies a filter to a list of objects. Arguments: (1) A list of objects, (2) A property to filter on, (3) A comparison operator, (4) A value to compare against. If property is boolean (unary), arguments: (1) A list of objects, (2) Unary property to filter on. Returns: A list of objects that pass the filter.
getProperty  # retrieves a property from an object. Arguments: (1) An object, (2) A property name. Returns: The value of the property.
reverse  # reverses the direction of a property. Arguments: (1) A property name. Returns: The reversed property name.
singleton  # creates a singleton set containing a single object. Arguments: (1) An object. Returns: A singleton set containing the object.
domain  # retrieves the domain of a property, which is the set of entities or objects that the property can be applied to. Arguments: (1) A property name. Returns: The set of entities that can have the property.
countSuperlative  # finds the object(s) with the minimum or maximum count of a certain property. Arguments: (1) A list of objects, (2) A superlative operator (min or max), (3) A property to count, (4) A list of objects to count from. Returns: The object(s) with the minimum or maximum count of the property.
ensureNumericProperty  # ensures that a property is treated as numeric for comparison purposes. Arguments: (1) A property name. Returns: The property name, treated as numeric.
ensureNumericEntity  # ensures that an entity is treated as numeric for comparison purposes. Arguments: (1) An entity. Returns: The entity, treated as numeric.
size  # retrieves the size of a collection. Arguments: (1) A collection of objects. Returns: The size of the collection as a numeric value.
aggregate  # applies an aggregate function to a property over a set of objects. Arguments: (1) An aggregate function (e.g., sum, avg, min, max), (2) A property to aggregate over, (3) A set of objects. Returns: The result of the aggregate function.
concat  # concatenates two or more strings or lists. Arguments: (1 and subsequent) Strings or lists to concatenate. Returns: The concatenated result.
countComparative  # compares the count of a property over a set of objects with a given number. Arguments: (1) A set of objects, (2) A property to count, (3) A comparison operator, (4) A number to compare against, (5) A set of objects to count from. Returns: The objects for which the count of the property satisfies the comparison.
superlative  # finds the object(s) with the minimum or maximum value of a certain property. Arguments: (1) A set of objects, (2) A superlative operator (min or max), (3) A property to compare. Returns: The object(s) with the minimum or maximum value of the property.

lambda  # creates a function. Arguments: (1) A variable name, (2) A function body. Returns: A function.
var  # references a variable. Arguments: (1) A variable name. Returns: The value of the variable.

# The following are namespaces for different types of entities.
en.person
en.company
en.university
en.relationship_status
en.employee
en.student
en.field
en.city
en.gender

# specific entities under these namespaces:
en.gender.female
en.gender.male
en.relationship_status
en.relationship_status.single
en.relationship_status.married

# en.person properties:
height  # property of type (number with unit en.cm)
birthdate  # property of type date
birthplace  # property of type en.city
logged_in  # property of type bool
friend  # property of type en.person
relationship_status  # property of type en.relationship_status

# education properties:
student  # property of type en.person
university  # property of type en.university
field_of_study  # property of type en.field
education_start_date  # property of type date
education_end_date  # property of type date

# employment properties:
employee  # property of type en.person
employer  # property of type en.company
job_title  # property of type string
employment_start_date  # property of type date
employment_end_date  # property of type date
\end{minted}
    \caption{Domain description for Overnight, using $\lambda$-DCS Simple..
    }
    \label{fig:overnight-lambda-dcs-simple}
\end{figure*}

\begin{figure*}
    \centering
\begin{minted}{text}
```javascript
const Gender = Object.freeze({"male":1, "female":2});
const RelationshipStatus = Object.freeze({"single":1, "married":2});

class Education {
    constructor(university, field_of_study, start_date, end_date) {
        this.university = university;
        this.field_of_study = field_of_study;
        this.start_date = start_date;
        this.end_date = end_date;
    }
}

class Employment {
    constructor(employer, job_title, start_date, end_date) {
        this.employer = employer;
        this.job_title = job_title;
        this.start_date = start_date;
        this.end_date = end_date;
    }
}

class Person {
    constructor(name, gender, relationship_status, height, birthdate, birthplace, friends = [], logged_in = false, education = [], employment = []) {
        this.name = name;
        this.gender = gender;
        this.relationship_status = relationship_status;
        this.height = height;
        this.birthdate = birthdate;
        this.birthplace = birthplace;
        this.friends = friends;
        this.logged_in = logged_in;
        this.education = education;
        this.employment = employment;
    }
}

class API {
    constructor(people = []) {
        this.people = people;
    }

    find_person_by_id(block_id) { ... }
}

let api = new API();
```
\end{minted}
    \caption{Domain description for Overnight, using Javascript. Continued from Fig.~\ref{fig:smcalflow-js}.
    }
    \label{fig:overnight-js}
\end{figure*}

\begin{figure*}
    \centering
\begin{minted}{text}
```scala
object Gender extends Enumeration {
  type Gender = Value
  val Male, Female = Value
}

object RelationshipStatus extends Enumeration {
  type RelationshipStatus = Value
  val Single, Married = Value
}

case class Education(university: String, fieldOfStudy: String, startDate: LocalDate, endDate: LocalDate)
case class Employment(employer: String, jobTitle: String, startDate: LocalDate, endDate: LocalDate)

case class Person(
  name: String,
  gender: Gender.Gender,
  relationshipStatus: RelationshipStatus.RelationshipStatus,
  height: Int,
  birthdate: LocalDate,
  birthplace: String,
  friends: Option[List[Person]] = None,
  loggedIn: Boolean = false,
  education: Option[List[Education]] = None,
  employment: Option[List[Employment]] = None
)

class API {
  var people: List[Person] = List()

  def findPersonById(personId: String): Person = ???
}

val api = new API
```
\end{minted}
    \caption{Domain description for Overnight, using Scala. 
    }
    \label{fig:overnight-sc}
\end{figure*}

\endgroup

\section{Prompt Construction}
\label{app:prompt}

We provide the prompt template that we use in Fig.~\ref{fig:prompt-template}.

% \begingroup
% \setlength{\tabcolsep}{3pt} % Default value: 6pt
% \renewcommand{\arraystretch}{0.8} % Default value: 1

\end{document}